%% file: Camera_Ready.tex
\documentclass{article} 
\usepackage[final,main]{neurips_2026}
\usepackage{hyperref}
\usepackage{url}
\usepackage{amssymb}
\usepackage{amsmath}
\usepackage{algorithm}
\usepackage{algorithmic}

\usepackage{enumitem}
\usepackage{booktabs}
\usepackage{graphicx}
\usepackage{multirow}
\usepackage{setspace}
\usepackage{makecell}
\usepackage{wrapfig}
\usepackage[table]{xcolor}
\usepackage{pifont}
\usepackage{enumitem}
\usepackage{marvosym}
\usepackage{bbm}
\usepackage[T1]{fontenc}
\usepackage{fontenc}   
\usepackage{soul}

\makeatletter
\def\@fnsymbol#1{%
  \ifcase#1
  \or *
  \or \text{\Letter}
  \or $\dagger$
  \or $\ddagger$
  \or $\mathsection$
  \or $\mathparagraph$
  \or $\|$
  \or $**$
  \or $\dagger\dagger$
  \or $\ddagger\ddagger$
  \else\@ctrerr
  \fi}
\makeatother
\title{Conformal Cache: Reliable Proxy-Discrepancy Caching for Fast Generative Inference}

\author{Wenxiao Wu$^{1,2,}$\thanks{Equal contributions.} \quad Yukang Feng$^{2,3,}$\footnotemark[1]  \quad Yachen Gao$^{2,4}$ \quad Chengming Xu$^{4}$\thanks{Co-corresponding authors.} \quad Moran Li$^{5}$ \\
\textbf{Jingyu Li}$^{2,4}$ \quad \textbf{Ming Xie}$^{2,4}$ \quad \textbf{Xiaobin Hu}$^{6}$  
\quad \textbf{Xinwei Sun}$^{4}$ \quad \textbf{Jing-Hao Xue}$^{7}$ \\  \textbf{Nong Sang}$^{1}$\footnotemark[2] \quad \textbf{Yanwei Fu$^{2,4}$} \\
\textsuperscript{1} Huazhong University of Science and Technology \quad 
\textsuperscript{2} Shanghai Innovation Institute \\
\textsuperscript{3} Nankai University \quad \textsuperscript{4} Fudan University 
\quad
\textsuperscript{5} ByteDance \\
\textsuperscript{6} National University of Singapore \quad \textsuperscript{7} University College London \\
\texttt{\{wenxiaowu, nsang\}@hust.edu.cn}\quad \texttt{yukangfeng@nankai.edu.cn} \\
\texttt{moramli.aca@gmail.com} \quad \texttt{xbhunanu@126.com} \quad \texttt{jinghao.xue@ucl.ac.uk}\\
\texttt{\{ycgao23, cmxu18, jingyuli24, mxie20, sunxinwei, yanweifu\}@fudan.edu.cn}
}

\newcommand{\methodname}{CCache }

\begin{document}

\maketitle
\begin{abstract}
While diffusion and flow-based generative models have achieved impressive performance in high-fidelity image and video synthesis, this capability comes at the cost of substantial inference overhead during iterative sampling.
Cache-based acceleration alleviates this cost by reusing intermediate computations across adjacent denoising or transport steps, typically relying on lightweight proxy discrepancies to decide when cached outputs should be refreshed. 
However, such proxy-discrepancy rules are inherently unreliable: the proxy is only a point estimate of the inaccessible oracle discrepancy and can misalign across prompts, timesteps, and generation dynamics. 
This mismatch induces asymmetric failures: proxy underestimation may cause unsafe reuse of stale cached computations and degrade generation quality, while proxy overestimation leads to overly conservative refreshes and reduced acceleration. 
To address these issues, this paper introduces Conformal Cache, dubbed CCache, a training-free conformal calibration framework for reliable cache-based generation acceleration. 
Our key insight is to formulate cache reuse as a one-sided risk-control problem and adaptively calibrate a timestep-wise upper correction for the signed proxy-to-oracle gap based on split conformal prediction. 
CCache is plug-and-play and can be integrated with different proxy-discrepancy caching baselines without structural modifications or parameter updates to the generative model. 
Extensive experiments on image and video generation demonstrate that \methodname improves generation quality and cache reliability while preserving favorable speed-quality trade-offs.
The source code is available in \href{https://github.com/Wu-Wenxiao/CCache}{https://github.com/Wu-Wenxiao/CCache}.
\end{abstract}

\section{Introduction}

Diffusion models and flow-based generative models \cite{flux2024, wan2025wan, hong2022cogvideo} have recently achieved remarkable progress in high-fidelity image and video generation. 
Their strong generative capability, however, comes at the cost of substantial inference overhead, as producing a single sample typically requires repeatedly performing full forward passes through a large neural network across many denoising or transport steps.
%
%
Consequently, reducing the number of expensive network evaluations without fine-tuning the model has emerged as a critical problem for practical deployment.

A prominent line of recent work \cite{ma2024deepcache,wimbauer2024cache,zhao2024pab} accelerates diffusion inference through cache-based computation reuse. 
These methods exploit the strong temporal redundancy between adjacent denoising or transport steps, allowing intermediate features, attention outputs, residuals, or other model outputs from previous steps.
Since directly verifying the safety of cache reuse would require additional full network evaluations, existing methods \cite{liu2024teacache, bu2025dicache} usually rely on lightweight proxy discrepancies to determine cache refreshes.
These proxies may be instantiated as timestep-modulated input differences, shallow-layer feature variations, transformation-vector changes, or residual-magnitude estimates.
Such proxy-discrepancy caching methods are appealing because they are training-free, efficient, and compatible with any pretrained generative backbones.

\begin{figure}[tb!]
\footnotesize
\centering

\scalebox{0.95}{
\includegraphics[width=1\linewidth]{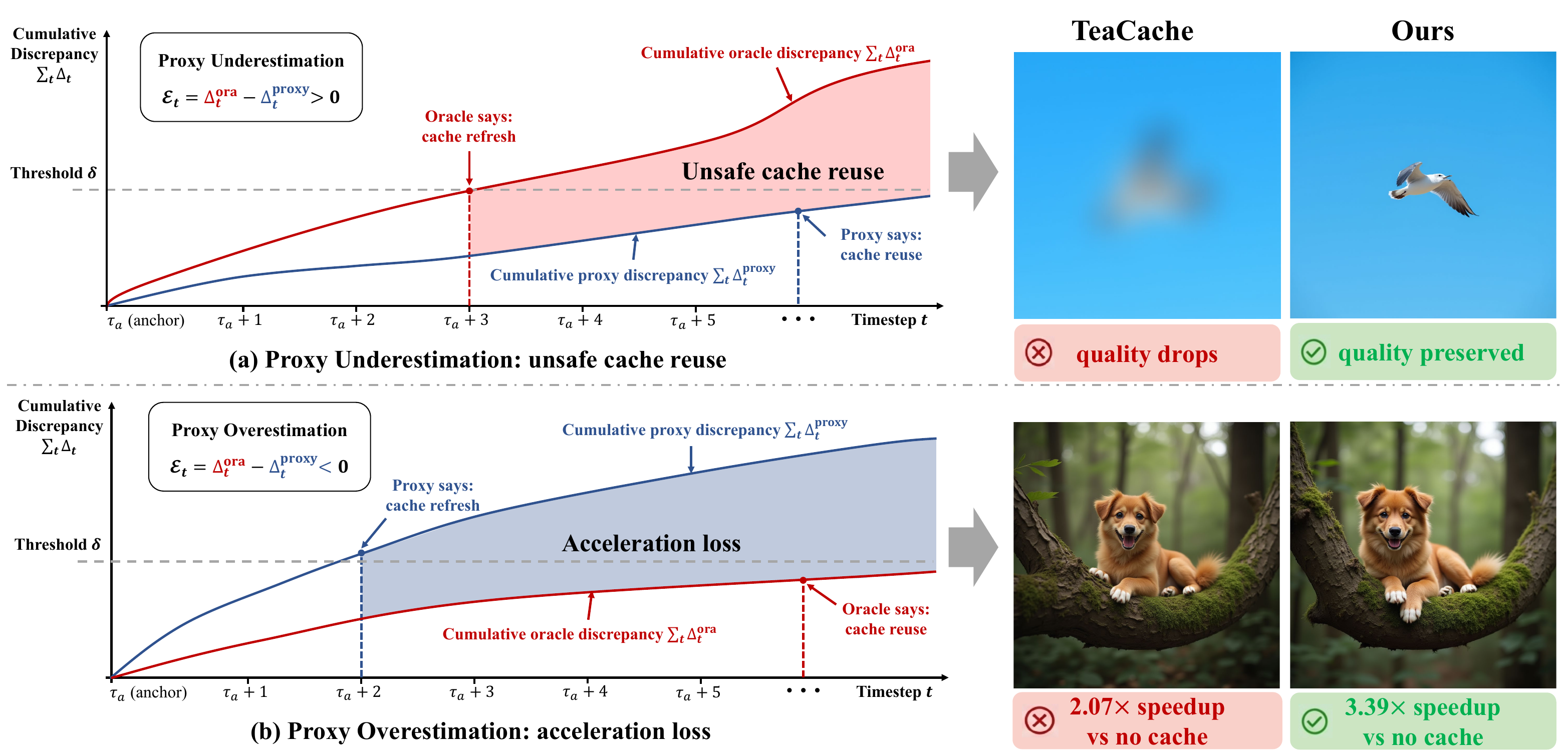} 
}
\vspace{-0.5em}
\caption{Asymmetric failure modes of proxy-discrepancy caching. These asymmetric consequences motivate CCache to prioritize a reliable upper correction for the signed proxy-to-oracle gap, preventing unsafe underestimation while avoiding excessive conservatism.}
\label{fig:teaser}
\vspace{-2em}
\end{figure}

Despite their empirical success, proxy-discrepancy cache rules suffer from a fundamental reliability issue: the cheap proxy is only a point estimate of the inaccessible oracle discrepancy. 
Since proxy-to-oracle alignment inevitably shifts across prompts, timesteps, and generation dynamics, the proxy cannot remain uniformly accurate in practice.
This unavoidable mismatch gives rise to two asymmetric failure modes:
\textbf{(1) \textit{The proxy underestimation.}}
When the proxy underestimates the oracle discrepancy, the model may falsely regard stale cached computations as still reliable.
As demonstrated in Figure \ref{fig:teaser}(a), cache reuse can continue even after the true oracle discrepancy has exceeded the tolerated threshold.
Since generative inference is inherently iterative, such uncorrected errors may propagate and accumulate across subsequent steps, eventually leading to visible artifacts, temporal inconsistency, or even generation collapse.
Therefore, reliable cache acceleration requires not only estimating oracle discrepancy, but also controlling the unsafe one-sided underestimation risk.
\textbf{(2) \textit{The proxy overestimation.}}
When the proxy overestimates the oracle discrepancy, the cache-refresh condition is violated earlier than necessary.
This behavior is generally safe from the perspective of generation quality, because stale cached computations are not overused.
However, as shown in Figuer \ref{fig:teaser}(b), it reduces the achievable acceleration ratio by triggering redundant full network evaluations.
Therefore, robust proxy-discrepancy caching should prioritize a trustworthy upper bound on the oracle discrepancy to prevent unsafe underestimation, while avoiding excessive conservatism that unnecessarily sacrifices acceleration.
%

To this end, we propose \textbf{C}onformal \textbf{C}ache, dubbed \textbf{CCache}, a calibration framework that adaptively upper-bounds proxy estimation risk for robust cache-based generation acceleration.
Our key observation is that cache-refresh decisions can be formulated as a one-sided risk-control problem.
This motivates calibrating an upper correction for the signed proxy-to-oracle gap, which directly targets the unsafe underestimation tail.
Built on this principle, CCache addresses the two failure modes discussed above through the following major improvements:
(1) \textit{Timestep-wise Conformal Correction for Safe Cache Reuse. }
%
Given a fixed proxy-discrepancy function inherited from an existing cache baseline, \methodname uses a small calibration set to run full inference and measure the residual between the oracle discrepancy and the proxy discrepancy at each timestep. 
Split conformal prediction is then used to estimate timestep-wise upper quantiles of this signed residual. 
During inference, these calibrated corrections are added to the original proxy discrepancies before applying the cache-refresh rule. 
In this way, \methodname transforms heuristic proxy reuse into calibrated proxy-discrepancy reuse, reducing the risk of unsafe underestimation without modifying or retraining the generative model.
(2) \textit{Online Miscoverage Adaptation for Efficient Cache Reuse. }
To mitigate the conservatism and efficiency degradation caused by fixed high-coverage correction, we further introduce an online segment-level miscoverage adaptation strategy.
Whenever the cache is refreshed, the full model outputs at the anchor and refresh timesteps are available, allowing \methodname to assess whether the previous cache segment has exceeded the tolerated oracle discrepancy.
%
%
If a violation is observed, \methodname decreases the miscoverage level to produce a more conservative correction; otherwise, it increases the miscoverage level to relax the correction and encourage more cache reuse.
This adaptive strategy improves acceleration under reliable cache reuse, while avoiding the excessive conservatism of a fixed high-coverage setting.
%
%
Above all, CCache is fully training-free and broadly applicable.
It only calibrates the residual behavior of an existing proxy and inserts the resulting correction into the cache-refresh criterion. 
Therefore, CCache can serve as a plug-in reliability layer for different proxy-discrepancy caching methods.
%
%

We summarize our contributions as four-folds:
{
(1) We introduce a new formulation of proxy-discrepancy caching as a one-side reliability problem, 
}
showing that proxy underestimation can induce cache-related generation degradation, whereas proxy overestimation causes conservative refreshes and reduced acceleration;
(2) We propose a split conformal prediction-based correction strategy to calibrate the signed proxy-to-oracle gap, thereby transforming heuristic proxy reuse into reliable cache reuse
{
. The resulting method-agnostic, plug-and-play reliability layer accommodates different proxy mechanisms without modifying or retraining the generative model;
}
(3) We introduce an online adaptive miscoverage strategy to mitigate the conservatism of fixed high-coverage correction, dynamically balancing cache reliability and acceleration efficiency;
(4) Extensive experiments {across methods, generation tasks, and model scales} demonstrate that our \methodname can substantially improve the robustness of proxy-discrepancy caching methods, achieving better generation quality and cache reliability while preserving favorable speed-quality trade-offs.

    
\vspace{-0.5em}
\section{Related work}
\vspace{-0.5em}
\textbf{Cache-based Acceleration for Generative Models and Proxy-Discrepancy Caching.}
Cache-based acceleration methods exploit temporal redundancy in diffusion inference by reusing intermediate computations across adjacent denoising steps, often without modifying or retraining the pretrained backbone.
DeepCache~\cite{ma2024deepcache} reuses cached high-level features while updating low-level ones cheaply.
Block Caching~\cite{wimbauer2024cache} extends this idea to layer-wise reuse by caching block outputs according to their temporal variation.
For video diffusion transformers, Pyramid Attention Broadcast ~\cite{zhao2024pab} broadcasts attention outputs across timesteps by exploiting their U-shaped temporal variation.
Beyond architecture-specific reuse, recent methods increasingly adopt proxy-discrepancy rules, where lightweight proxy signals are used to determine when to refresh cached computations.
Among these methods, TeaCache~\cite{liu2024teacache} estimates model-output differences from timestep-modulated input differences.
EasyCache~\cite{zhou2025easycache} dynamically reuses previously computed transformation vectors without offline profiling or pre-computation,
and MagCache~\cite{ma2025magcache} exploits a magnitude law for adaptive timestep skipping.
DiCache~\cite{bu2025dicache} uses shallow-layer probes for sample-specific cache decisions.
Beyond direct reuse, GOC~\cite{qiu2025accelerating} compensates
caching errors using temporal feature gradients and inflection-aware
correction.
TaylorSeer~\cite{liu2025reusing} forecasts future features through
higher-order Taylor expansions.
ERTACache~\cite{peng2026ertacache} combines residual error
rectification with trajectory-aware timestep adjustment to mitigate
feature shift and step amplification.
%
Rather than improving feature approximations or designing a new
discrepancy proxy, we revisit proxy-discrepancy caching from a reliability perspective and formulate cache reuse as a one-sided risk-control problem.
%
%
By directly controlling the unsafe underestimation tail, CCache improves the reliability of cache reuse.
%


\textbf{Conformal Prediction and Adaptive Conformal Inference.}
Conformal prediction~\cite{shafer2008tutorial,zhou2024conformal,sun2022conformal} provides distribution-free uncertainty quantification with theoretically guaranteed coverage probabilities under exchangeability.
Conformal prediction has attracted increasing attention in tasks that require rigorous uncertainty quantification, including multi-class prediction~\cite{ding2023class}, In-context learning~\cite{wu2026towards}, LLM benchmarking~\cite{ye2024benchmarking}, and robotic trajectory prediction~\cite{sun2023conformal}.
Split conformal prediction (SCP)~\cite{lei2018distribution,oliveira2024split} is especially relevant to our setting because it calibrates a fixed prediction rule on held-out data without retraining the underlying model.
Extensions further study split conformal prediction under certain non-exchangeable processes via explicit coverage penalties~\cite{oliveira2024split}.
Beyond static calibration, adaptive conformal inference updates the miscoverage level online to handle distribution shifts over time~\cite{gibbs2021adaptive}.
It is also related to conformal risk-control frameworks, which calibrate black-box predictors to satisfy user-specified finite-sample risk bounds~\cite{bates2021distribution}.
Motivated by adaptive conformal inference, CCache dynamically updates the miscoverage level of the SCP-calibrated upper correction using an online cache-reliability signal, effectively reducing excessive conservatism while maintaining reliable cache reuse.
\vspace{-0.5em}
\section{Methodology}
\vspace{-0.5em}
\subsection{Preliminary: a general formulation of Proxy-Discrepancy Caching}
\textbf{Oracle Discrepancy and Ideal Caching Condition.} In diffusion models and flow matching, generation is performed by discretizing a continuous denoising or transport trajectory into $T$ timesteps.
At each timestep $t$, the model evaluates a full neural network $\Phi$ to obtain the velocity, residual, or feature output. 
We denote the oracle output at timestep $t$ as
$\mathbf{O}_t = \Phi(\mathbf{x}_t, t, c),$
where $\mathbf{x}_t$ is the current latent state and $c$ denotes the conditioning signal like a text prompt. 
%
Since evaluating $\Phi$ at every timestep dominates the inference cost, caching-based acceleration methods aim to skip redundant full forward passes by reusing a cached output from a previous anchor timestep $t_a$.
%
%
Given a distance function $d(\cdot,\cdot)$, such as relative $L_1$ distance, the oracle discrepancy between consecutive timesteps is defined as $d(\mathbf{O}_t, \mathbf{O}_{t+1})$.
Starting from an anchor timestep $\tau_a$, cached computation can be safely reused as long as the cumulative oracle discrepancy remains below a predefined threshold $\delta$:
\begin{equation}
    \sum_{t=\tau_a}^{\tau-1} d(\mathbf{O}_t, \mathbf{O}_{t+1}) \le \delta.
\end{equation}
%
%
Once the cumulative discrepancy exceeds $\delta$, the cache should be refreshed by performing a full network evaluation and resetting the anchor.
However, directly computing $d(\mathbf{O}_t,\mathbf{O}_{t+1})$ requires evaluating $\Phi$ at both timesteps, which defeats the purpose of caching. 

\textbf{Proxy-Discrepancy Function. }Proxy-Discrepancy Caching introduces a lightweight proxy function $g(\cdot,\cdot)$ to estimate the oracle discrepancy without running the full network. 
Since the oracle discrepancy is defined between consecutive model outputs, the proxy is generally computed from the corresponding low-cost feature variation between adjacent timesteps. 
%
Let $\mathbf{F}_t$ denote an easily accessible feature at timestep $\tau$. 
The proxy estimates the single-step oracle discrepancy as
\begin{equation}
    g(\mathbf{F}_t, \mathbf{F}_{t+1})
\approx
d(\mathbf{O}_t, \mathbf{O}_{t+1}).
\end{equation}
%
Different caching methods instantiate $g$ in different ways. 
%
For example, TeaCache~\cite{liu2024teacache} uses a polynomially corrected difference of timestep-modulated inputs, while DiCache uses shallow-layer feature variation as an online proxy for output variation.
Despite using different proxy signals, both methods share the same underlying principle: replacing the expensive oracle discrepancy with a cheap error predictor.

During inference, Proxy-Discrepancy Caching avoids full network evaluation at every timestep. 
Instead, it accumulates the predicted discrepancy given by the proxy function and reuses the cache while 
$\sum_{t=\tau_a}^{\tau-1} g(\mathbf{F}_t, \mathbf{F}_{t+1}) \le \delta$.
A cache refresh is triggered at the first timestep $\tau_b$ where
\begin{equation}
    \sum_{t=\tau_a}^{\tau_b-1} g(\mathbf{F}_t, \mathbf{F}_{t+1}) \le \delta < \sum_{t=\tau_a}^{\tau_b} g(\mathbf{F}_t, \mathbf{F}_{t+1}).
\end{equation}
At this point, the model performs a full forward pass, updates the cached output, clears the accumulated error, and resets the anchor as $\tau_a \leftarrow \tau_b$.

\textbf{Motivation: The Proxy-to-Oracle Gap.}
Although Proxy-Discrepancy Caching provides a general and efficient formulation for generation acceleration, it also exposes a fundamental limitation. 
The proxy $g$ is typically a deterministic point estimate calibrated from observed data. 
Therefore, it may fail to precisely predict the true oracle discrepancy under diverse prompts, complex motion patterns, rare visual concepts, or out-of-distribution inputs.
We define the proxy-to-oracle gap as
\begin{equation}
\label{eq:gap}
 \mathcal{E}_t = d(\mathbf{O}_{t-1}, \mathbf{O}_{t}) - g(\mathbf{F}_{t-1}, \mathbf{F}_{t}).
\end{equation}

\textbf{Analysis.} This generalization failure can manifest in two directions, creating highly asymmetric consequences that demand distinct strategies:
\vspace{-0.5em}
\begin{itemize}[leftmargin=8pt]
    \item \textbf{Severe Underestimation ($\mathcal{E}_t \gg 0$):} The proxy substantially underestimates the oracle discrepancy, causing the model to falsely regard the cached output as still reliable. 
    Consequently, The accumulated proxy discrepancy may remain below $\delta$ even after the true oracle discrepancy has exceeded the tolerated range, leading to prolonged cache reuse. 
    Since the reverse generative trajectory is iterative, such uncorrected deviations can propagate and amplify across successive timesteps, eventually leading to even catastrophic generation collapse.
    This motivates moving beyond heuristic point estimation for robust quality preservation.
    \textit{To safely constrain this risk, we establish a reliable upper bound on $\mathcal{E}_\tau$ using Split Conformal Prediction (detailed in Section \ref{sec:3.2}).}

    \item \textbf{Conservative Overestimation ($\mathcal{E}_t < 0$):} 
    The proxy overestimates the oracle discrepancy, causing the caching condition to be violated prematurely. 
    While this is generally safe and preserves generation quality, it leads to more frequent cache refreshes and thus a suboptimal acceleration ratio.
    %
    %
    \textit{To recover efficiency after conservative correction, we further adapt the calibration miscoverage level online using a segment-level reliability signal (detailed in Section \ref{sec:3.3}))}.
    
\end{itemize}
%
\subsection{Timestep-wise conformal correction for safe cache reuse}
\label{sec:3.2}
This section details the construction of a reliable timestep-wise correction for Proxy-Discrepancy Caching, which upper-bounds the proxy-to-oracle gap $\mathcal{E}_t$ with a user-specified significance level \(\alpha\), or equivalently a coverage level \(1-\alpha\), guided by the theory of split conformal prediction.

We begin by randomly sampling a minuscule calibration subset of prompts from the target evaluation dataset $\mathcal{X}_{test}$, denoted as $\mathcal{X}_{cal}=\{c_i\}_{i=1}^{N}\subset\mathcal{X}_{test}$, where $N\ll |\mathcal{X}_{test}|$.
The remaining prompts \(\mathcal{X}_{eval}=\mathcal{X}_{test}\setminus\mathcal{X}_{cal}\) are reserved for evaluation.
Please note that we do not need extra data to train a predictive model, because the proxy-discrepancy function \(g(\cdot,\cdot)\) is inherited from existing caching baselines, such as the \textit{poly1d }function from TeaCache~\cite{liu2024teacache}, and is used as a fixed discrepancy estimator.
\(\mathcal{X}_{cal}\) is only used to calibrate the proxy-to-oracle gap defined in Eq.~(\ref{eq:gap}).
For each prompt \(c_i\in\mathcal{X}_{cal}\), we run the full \(T\)-step generation process without caching.
This gives the oracle outputs \(\{\mathbf{O}_{i,t}\}_{t=1}^{T}\) and the corresponding lightweight proxy features \(\{\mathbf{F}_{i,t}\}_{t=1}^{T}\).
For each timestep \(t\in\{2,\ldots,T\}\), we compute the oracle discrepancy and proxy discrepancy between the current timestep and its previous timestep as
\begin{equation}
    \Delta^{\mathrm{ora}}_{i,t}
    =
    d(\mathbf{O}_{i,t-1},\mathbf{O}_{i,t}),
    \qquad
    {\Delta}_{i,t}^{\mathrm{proxy}}
    =
    g(\mathbf{F}_{i,t-1},\mathbf{F}_{i,t}).
\end{equation}
Based on the above definitions, the unconformity score \(s_{i,t}\) for each calibration prompt \(c_i\) at timestep \(t\) is calculated as the proxy-to-oracle gap:
\begin{equation}
\label{eq:sample_gap}
    \mathcal{E}_{i,t}
    =
    \Delta^{\mathrm{ora}}_{i,t}
    -
    \Delta^{\mathrm{proxy}}_{i,t}.
\end{equation}
We use the signed proxy-to-oracle gap \(\mathcal{E}_{i,t}\), rather than its absolute value, as the nonconformity score because proxy errors have asymmetric consequences in cache reuse.
A positive gap indicates proxy underestimation, which is unsafe since stale cached outputs may be reused beyond the tolerated range.
In contrast, a negative gap indicates proxy overestimation, which is conservative buys safe, and mainly leads to earlier refreshes and reduced acceleration.
Therefore, instead of pursuing a two-sided accurate proxy, our objective is to prioritize a reliable upper correction for the unsafe underestimation direction.
%
%
%
This procedure is repeated for every \(c_i\in\mathcal{X}_{cal}\) at each timestep \(t\).
Then, for each timestep \(t\), a set including all \(N\) unconformity scores can be denoted by
\begin{equation}
\label{eq:timestep_unconformity_set}
    \mathcal{S}_t
    =
    \{\mathcal{E}_{i,t}\}_{i=1}^{N}
    \cup
    \{+\infty\}.
\end{equation}
Subsequently, the corresponding upper \((1-\alpha)\)-quantile \(q_{t,1-\alpha}\), which serves as the timestep-specific reliability correction, is determined from the empirical distribution of unconformity scores in \(\mathcal{S}_t\):
\begin{equation}
\label{eq:timestep_conformal_quantile}
    q_{t,1-\alpha}
    =
    \text{the }
    \left\lceil (1-\alpha)(N+1) \right\rceil
    \text{-th smallest value in } \mathcal{S}_t .
\end{equation}
%
%
Under the exchangeability assumption between calibration and evaluation prompts, split conformal prediction guarantees that, for each timestep \(t\),
\begin{equation}
\label{eq:cp_gap_guarantee}
    \mathbb{P}
    \left[
        \mathcal{E}_{t}
        \le
        q_{t,1-\alpha}
    \right]=\mathbb{P}
    \left[
        \Delta^{\mathrm{ora}}_{t}
        \le
        \Delta^{\mathrm{proxy}}_{t}
        +
        q_{t,1-\alpha}
    \right]
    \ge
    1-\alpha .
\end{equation}
%
%
This conformal guarantee is timestep-wise and marginal, and thus should not be interpreted as a formal trajectory-level coverage guarantee.
Therefore, by adding the conformal correction \(q_{t,1-\alpha}\) to the original proxy discrepancy, we obtain a high-confidence upper estimate of the oracle discrepancy:
\begin{equation}
\label{eq:corrected_proxy}
    \widetilde{\Delta}^{\mathrm{proxy}}_{t}
    =
    \Delta^{\mathrm{proxy}}_{t}
    +
    q_{t,1-\alpha}.
\end{equation}
After calibration, all timestep-wise corrections constitute a vector
\begin{equation}
\label{eq:correction_vector}
    \mathbf{q}_{1-\alpha}
    =
    [q_{2,1-\alpha},q_{3,1-\alpha},\ldots,q_{T,1-\alpha}],
\end{equation}
which characterizes the timestep-dependent proxy-to-oracle gap along the generation trajectory.
During inference, for any prompt \(c\in\mathcal{X}_{eval}\), we directly use this correction vector to replace the original proxy discrepancy \(\Delta^{\mathrm{proxy}}_{t}\) with its corrected version \(\widetilde{\Delta}^{\mathrm{proxy}}_{t}\) in the cache reuse criterion.
%
%
Specifically, starting from an anchor timestep \(\tau_a\), a cache refresh is triggered at the first timestep \(\tau_b\) such that
\begin{equation}
\label{eq:cp_caching_rule}
    \sum_{t=\tau_a+1}^{\tau_b-1}
    \left[
        g(\mathbf{F}_{t-1},\mathbf{F}_{t})
        +
        q_{t,1-\alpha}
    \right]
    \le
    \delta
    <
    \sum_{t=\tau_a+1}^{\tau_b}
    \left[
        g(\mathbf{F}_{t-1},\mathbf{F}_{t})
        +
        q_{t,1-\alpha}
    \right].
\end{equation}
%
%
Compared with the original proxy-only caching rule, our conformally corrected rule explicitly accounts for the calibrated proxy-to-oracle gap, thereby reducing the risk of severe proxy underestimation while retaining the efficiency of lightweight proxy accumulation.


\textbf{Remark.}
{The conformal correction provides a timestep-wise marginal upper bound on the proxy-to-oracle gap. Since the cache-refresh rule accumulates corrected discrepancies over a segment, CCache should be viewed as a conformally calibrated surrogate for safer cumulative cache. 
}

\subsection{{Online miscoverage adaptation for efficient cache reuse}}
\label{sec:3.3}
The conformal correction above effectively reduces the risk of unsafe proxy underestimation.
However, under the same caching threshold \(\delta\), adding \(q_{t,1-\alpha}\) to the original proxy discrepancy inevitably makes the cache-refresh rule more conservative than the original proxy-discrepancy caching method.
This effect becomes more pronounced when \(\alpha\) is small, since a smaller significance level corresponds to a larger coverage level \(1-\alpha\), and thus leads to a larger conformal correction.
Consequently, although a fixed high-coverage correction improves reliability, it may also trigger overly frequent cache refreshes and reduce the acceleration ratio.
Inspired by adaptive conformal inference, we further introduce an online adaptation strategy to better balance generation quality and inference efficiency, which updates the coverage level on the fly based on the observed discrepancy at cache refresh points.

Let \(\alpha_m\) denote the adaptive miscoverage level used for the \(m\)-th cache segment.
In the fixed conformal correction described above, \(\alpha_m\) remains unchanged throughout inference.
In contrast, our adaptive conformal inference updates \(\alpha_m\) online according to the observed reliability of each cache segment.
Specifically, given an anchor timestep \(\tau_a^m\), suppose the cache is refreshed at timestep \(\tau_b^m\).
Since full model outputs are available at both \(\tau_a^m\) and \(\tau_b^m\), we compute the observed oracle discrepancy as $D_m^{\mathrm{ora}} = d(\mathbf{O}_{\tau_a^m},\mathbf{O}_{\tau_b^m})$.
If \(D_m^{\mathrm{ora}}>\delta\), the true discrepancy has exceeded the tolerance threshold, indicating that the current correction is still too optimistic.
In this case, we decrease \(\alpha_m\), which increases the coverage level \(1-\alpha_m\) and yields a more conservative correction.
Otherwise, if \(D_m^{\mathrm{ora}}\le\delta\), the current correction is sufficiently safe, and we increase \(\alpha_m\) to reduce the correction and encourage more cache reuse.
Formally, we define the segment-level violation indicator as
\begin{equation}
\label{eq:aci_violation}
    v_m
    =
    \mathbbm{1}
    \left[
        D_m^{\mathrm{ora}}>\delta
    \right].
\end{equation}
Given a target miscoverage level \(\alpha_0\) and an update step size \(\eta\), we update the adaptive miscoverage level by
\begin{equation}
\label{eq:aci_update}
    \alpha_{m+1}
    =
    \Pi_{[\alpha_{\min},\alpha_{\max}]}
    \left(
        \alpha_m
        +
        \eta(\alpha_0-v_m)
    \right),
\end{equation}
where \(\Pi_{[\alpha_{\min},\alpha_{\max}]}\) denotes clipping to a predefined valid range.
When a violation occurs, i.e., \(v_m=1\), Eq.~(\ref{eq:aci_update}) decreases \(\alpha_m\), thereby increasing the coverage level \(1-\alpha_m\) and producing a larger conformal correction in the next segment.
When no violation occurs, i.e., \(v_m=0\), Eq.~(\ref{eq:aci_update}) increases \(\alpha_m\), thereby reducing the coverage level and allowing the method to reuse cached computation more aggressively.
In this way, Eq.~(\ref{eq:aci_update}) adaptively tightens or relaxes the correction according to the observed cache reliability, improving the acceleration ratio while preserving the safety benefit of conformal calibration.

When it comes to implementation, the calibration scores can be sorted once to support efficient quantile extraction.
For each candidate miscoverage level \(\alpha\), we directly extract the corresponding timestep-wise quantiles from the sorted calibration scores, yielding a correction vector \(\mathbf{q}_{1-\alpha}\).
During inference, the adaptive strategy simply selects the correction vector associated with the current \(\alpha_m\).
This lookup-based design incurs negligible additional overhead while enabling online adjustment of the conformal correction.

\section{Experiments}
\label{sec:4}

We give the setups, and evaluate CCache on two state-of-the-art proxy-discrepancy caching baselines, TeaCache~\cite{liu2024teacache} and DiCache~\cite{bu2025dicache}, across both image and video generation tasks.
\begin{table}[t]
    \centering
    \setlength{\tabcolsep}{2pt}
    \caption{Quantitative comparison on HPSV3 for text-to-image generation. ``+ CCache" denotes our calibrated version of each caching baseline. }
    \label{tab:t2v_1}
    \scalebox{0.71}{
    \begin{tabular}{c l| c c| c c c c c c}
        \toprule
\multirow{2}{*}{\textbf{Model}} & \multirow{2}{*}{\textbf{Method}} & \multirow{2}{*}{\textbf{Speedup}$\uparrow$} & \multirow{2}{*}{\textbf{Latency}$\downarrow$} &\multicolumn{6}{c}{\textbf{HPSv3}}                                                                                                            \tabularnewline
                               && &                                  & \textbf{Characters}     & \textbf{Arts}     & \textbf{Design}   & \textbf{Architecture} & \textbf{Animals}  & \textbf{Natural Scenery} \tabularnewline
        \midrule
        & Vanilla (100\% steps)     &1.00$\times$ &  6.45s      & -                  & -                  & -                  &           &              & - \\
        & Uniform Cache ($I=2$)     &1.90$\times$ &  3.40s     &  11.43            &    10.05          &       8.33        &   10.76        &    9.78          & 9.38 \\
        & TeaCache ($\delta=0.40$)   &1.98$\times$ &3.25s &11.55 &10.16 &8.93 &10.58 &10.05 & 9.34       \\
        
        \rowcolor{gray!25} \cellcolor{white} 
        
        & + CCache                  &2.05$\times$ &3.14s &11.59 &10.13 &9.04 &10.57 &10.14 &9.50  \\
        & TeaCache ($\delta=0.80$)   &2.99$\times$ &  2.16s      & 10.20 & 9.02         &7.57    & 9.36              & 8.48          &     7.71       \\
        
        \rowcolor{gray!25} \cellcolor{white} 
        
        & + CCache                  &3.14$\times$ & 2.06s       & 11.08 & 9.20             & 8.35              & 9.88          & 9.70            & 8.54 \\
        & DiCache ($\delta=0.40$)    & 2.82$\times$ &   2.29s     & 11.50 & 9.72 & 8.73 & 10.91 & 10.62 & 10.00 \\
        \rowcolor{gray!25} \cellcolor{white}
       \multirow{-8}{*}{\rotatebox{90}{\textbf{FLUX.1-dev}}} & +CCache      & 2.78$\times$ &   2.32s     & 11.91 & 10.06 & 9.01 & 11.47 & 11.15 & 10.39 \\
       \midrule
\multirow{2}{*}{\textbf{Model}} & \multirow{2}{*}{\textbf{Method}} & \multirow{2}{*}{\textbf{Speedup}$\uparrow$} & \multirow{2}{*}{\textbf{Latency}$\downarrow$} &\multicolumn{6}{c}{\textbf{HPSv3}}                                                                                                            \tabularnewline
                               && &                                  & \textbf{Transportation} & \textbf{Products} & \textbf{Others}   & \textbf{Plants}       & \textbf{Food}     & \textbf{Science} \tabularnewline
        \midrule
        & Vanilla (100\% steps)     &1.00$\times$ &  6.45s      & -                  & -                  & -                  &           &              & - \\
        & Uniform Cache ($I=2$)     &1.90$\times$ &  3.40s       &   10.58          &   10.05           &       9.75      &   8.99       & 10.41             & 8.30 \\
        & TeaCache ($\delta=0.40$)   &1.98$\times$ &3.25s &10.46 &10.10 &9.80 &9.56 &10.64 & 8.68     \\
        
        \rowcolor{gray!25} \cellcolor{white} 
        
        & + CCache                  &2.05$\times$ &3.14s &10.60 &10.30 &9.95 &10.03 &10.96 &8.77  \\
        & TeaCache ($\delta=0.80$)   & 2.99$\times$&  2.16s      & 8.95 & 8.53            & 8.45              & 6.46         & 8.61           & 7.26 \\
        
        \rowcolor{gray!25} \cellcolor{white} 
        
        & + CCache                  &3.14$\times$ & 2.06s       & 10.23 & 9.92             & 9.48              & 9.45          & 10.81            & 7.89 \\
        & DiCache ($\delta=0.40$)     & 2.82$\times$ &   2.29s        & 10.91 & 10.80 & 10.37 & 10.48 & 11.90 & 8.57 \\
        \rowcolor{gray!25} \cellcolor{white}
       \multirow{-8}{*}{\rotatebox{90}{\textbf{FLUX.1-dev}}} & + CCache & 2.78$\times$ & 2.32s & 11.56 & 11.39 & 10.85 & 11.03 & 12.33 & 8.93 \\
        \bottomrule
    \end{tabular}
    }
    \vspace{-0.5em}
\end{table}

\textbf{Experimental Settings.}
We conduct experiments on Wan2.1-1.3B~\cite{wan2025wan} and Wan2.1-14B~\cite{wan2025wan} for video generation, and FLUX.1-dev~\cite{flux2024} for image generation.
Since the conformal upper correction makes the CCache-calibrated rule more conservative under the same cache threshold, directly using the same threshold would typically lead to lower speedup.
For a fair comparison, we adjust the caching threshold of each CCache-enhanced variant to achieve a comparable acceleration ratio to its corresponding baseline.
Since reliability is our primary objective, we fix the target miscoverage level to $\alpha_0=0.05$. We use $N=100$ calibration samples, yielding an empirical quantile resolution of approximately $1/(N+1)\approx0.01$. Accordingly, we set $\eta=0.2$, with $\alpha_{min}=0.05$ and $\alpha_{max}=0.5$. Under the proposed update rule, each non-violating segment increases the miscoverage level by $0.01$, whereas a violating segment decreases it by $0.19$. Therefore, the correction is relaxed at a practically meaningful resolution but rapidly returns toward the conservative 95\% coverage baseline once a violation is observed.
We evaluate generation quality and inference efficiency, reporting VMBench~\cite{ling2025vmbench} and HPSv3~\cite{ma2025hpsv3widespectrumhumanpreference} scores together with speedup and latency to characterize the overall speed-quality trade-off.
 All experiments are conducted on 8 NVIDIA H100 GPUs.
More experimental details can be seen in Appendix~\ref{app:exp_details}.
%

%

%

\textbf{Evaluation Benchmarks.}
For video generation, we employ VMBench~\cite{ling2025vmbench} to assess motion quality.
VMBench is a perception-aligned benchmark designed for evaluating video motion generation, with a prompt suite covering diverse motion patterns and six major dynamic scene dimensions.
Following VMBench, we report five perception-driven motion metrics: \textit{Perceptible Amplitude Score (PAS), Object Integrity Score (OIS), Temporal Coherence Score (TCS), Motion Smoothness Score (MSS), and Commonsense Adherence Score (CAS)}.
%
For image generation, we adopt HPSv3~\cite{ma2025hpsv3widespectrumhumanpreference} to evaluate synthesis quality.
HPSv3 is a human-preference-based image evaluation model designed to assess generated images in terms of perceptual quality and text-image alignment.
For benchmarking, HPSv3 samples 1,000 prompts from each of the 12 HPDv3 categories, resulting in 12,000 prompts.
We report the HPSv3 score as the image-quality metric.

\begin{table}[t]
    \centering
    \renewcommand{\arraystretch}{1.1} 
    \caption{Quantitative comparison on VMBench for text-to-video generation. ``+CCache" denotes our calibrated version of each caching baseline. 
    }
    \label{tab:t2v_2}
    \scalebox{0.82}{
    \begin{tabular}{c l| c c| c c c c c c}
        \toprule
        \multirow{2}{*}{\textbf{Model}} & \multirow{2}{*}{\textbf{Method}}&\multirow{2}{*}{\textbf{Speedup}$\uparrow$} & \multirow{2}{*}{\textbf{Latency}$\downarrow$}&\multicolumn{6}{c}{\textbf{VMBench }} \tabularnewline
 &&&&\textbf{PAS} &\textbf{OIS} &\textbf{TCS} & \textbf{MSS} & \textbf{CAS} & \textbf{Avg.}
 \\
        \midrule
        
        & Vanilla (100\% steps)     &1.00$\times$ &  95s      & -                  & -                  & -                  & -          & -             & - \\
        & Uniform Cache ($I=2$)     &2.11$\times$ &  45s       & 10.17             &  66.03            &  94.07            &    71.09       &  55.98            & 59.46 \\
        & TeaCache ($\delta=0.08$)  &2.07$\times$ & 46s       & 8.78 & 67.91             & 94.08              & 71.62          & 56.44            & 59.77 \\
        
        \rowcolor{gray!25} \cellcolor{white}
        & + CCache                  &2.07$\times$ & 46s       & 8.79 & 68.65             & 94.41              & 72.54          & 57.23            & 60.32 \\
        
        & TeaCache ($\delta=0.20$)   &3.52$\times$ &  27s      & 8.17             & 67.14             & 93.38              & 65.93         & 53.78             & 57.68 \\
        \rowcolor{gray!25} \cellcolor{white}

        & + CCache                  &3.39$\times$ &28s        & 8.42 & 68.39             & 94.06              & 68.99          & 55.36            & 59.04 \\
        & DiCache ($\delta=0.20$)   & 1.76$\times$ &  54s      &  13.46 &    66.71  &  94.42 & 72.16  & 52.87         & 59.92 \\
        
        \rowcolor{gray!15} \cellcolor{white} \multirow{-8}{*}{\footnotesize{\rotatebox{90}{\textbf{Wan2.1-1.3B}}}} 
        & + CCache                  & 1.73$\times$ &  55s      & 14.19 &   67.07  &  94.63  & 75.63 & 54.48            & 61.20 \\
        \midrule
        & Vanilla (100\% steps)     &1.00$\times$ & 1794s       & -                  & -                  & -                  & -          & -             & - \\
        & Uniform Cache ($I=2$)     &1.99$\times$ &  903s      & -             & -            & -              & -          & -              & - \\
        
        & TeaCache ($\delta=0.20$)   &2.06$\times$  &  872s      & 15.95 & 71.39             & 94.72              & 78.99          & 63.79            & 64.97 \\
        
        \rowcolor{gray!15} \cellcolor{white} \multirow{-4}{*}{\footnotesize{\rotatebox{90}{\textbf{Wan2.1-14B}}}} 
        & + CCache                  &2.05$\times$ &  874s      & 17.55 & 74.56             & 95.63              & 82.82          & 65.91            & 67.29 \\

        \bottomrule
    \end{tabular}
    }
    \vspace{-1em}
\end{table}

\subsection{Main results}

\begin{figure}[t]
\footnotesize
\centering
\includegraphics[width=1\linewidth]{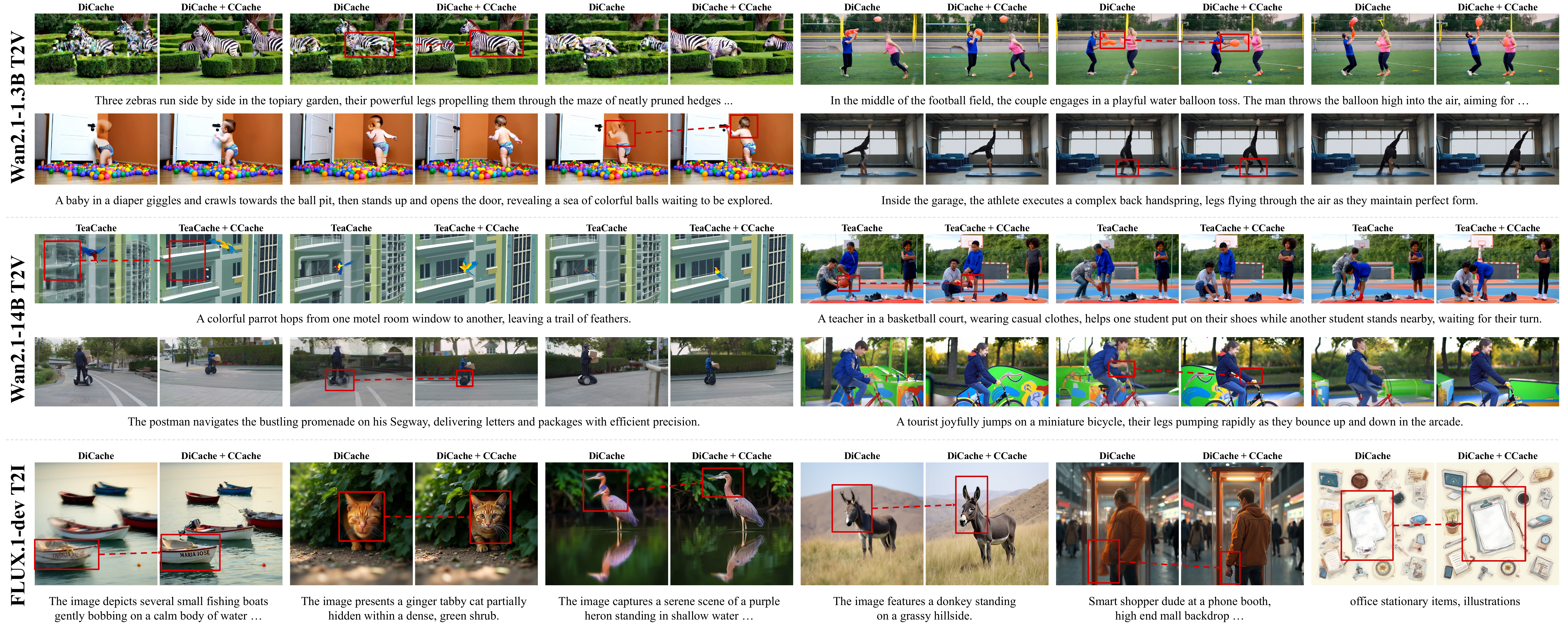}
\vspace{-2em}
\caption{
Qualitative comparison of baseline caching methods with and without CCache. Red boxes highlight representative differences. CCache reduces visible structural distortions, including merged subjects, malformed object boundaries, and loss of fine detail, under comparable acceleration.
}
\label{fig:qualitative_comparison}
\vspace{-2em}
\end{figure}

\textbf{Quantitative Analysis.}
The quantitative results for image and video generation are presented in Table~\ref{tab:t2v_1} and Table~\ref{tab:t2v_2}, respectively.
Here are several observations.
First, CCache consistently improves generation quality under comparable acceleration across both video and image generation tasks.
On Wan2.1-1.3B with TeaCache at \(\delta=0.08\), CCache maintains the same speedup and latency, i.e., \(2.07\times\) and 46s, while improving the average VMBench score from 59.77 to 60.32.
A consistent improvement is observed on FLUX.1-dev for image generation, demonstrating the generalization capability of CCache.
Second, the benefit of CCache becomes more pronounced under more aggressive caching.
On Wan2.1-1.3B, the average VMBench gain over TeaCache increases from +0.55 at \(\delta=0.08\) to +1.36 at \(\delta=0.20\).
Similarly, on image generation, the average HPSv3 improvement under TeaCache is larger at \(\delta=0.80\), where the score increases from 8.38 to 9.54.
This supports our motivation that \textit{conformal correction is especially beneficial when the baseline proxy is more likely to misestimate the oracle discrepancy}.
Third, the improvement is more substantial on the stronger Wan2.1-14B model.
At the same TeaCache setting \(\delta=0.20\), CCache improves the average VMBench score by +1.36 on Wan2.1-1.3B, while the gain increases to +2.32 on Wan2.1-14B.
This suggests that CCache can better preserve and exploit the stronger generative capability of larger models once unsafe proxy underestimation is constrained.

\textbf{Qualitative Analysis.}
Figure~\ref{fig:qualitative_comparison} shows that CCache mitigates localized structural artifacts in the outputs of the underlying caching methods. On Wan2.1-1.3B, DiCache produces visibly fragmented zebra bodies, discontinuous stripe patterns, and poorly defined legs. Adding CCache makes the individual animals easier to distinguish and preserves more coherent body contours across the sampled frames. In the other video examples, the improvement is evident around the infant’s face and arms, the water-balloon interaction, and the gymnast’s extended limbs, where errors in shape or articulation are particularly noticeable. On Wan2.1-14B, TeaCache distorts the window geometry surrounding the moving parrot, while CCache retains a more regular building structure. In the basketball-court example, the CCache result also depicts the requested shoe-fitting interaction more clearly, rather than obscuring it with an inconsistent object. The FLUX.1-dev examples exhibit a similar pattern: CCache yields better-defined boat hulls and stationery outlines, and avoids the conspicuous duplicated head in the heron image. Overall, these examples complement the quantitative results by illustrating that CCache improves perceptual quality across diverse scenes, model scales, and generation tasks, while maintaining acceleration comparable to the underlying caching methods. Additional qualitative results are provided in Appendix~\ref{appd:more_qualitative_results}.
\subsection{Ablation study}
\textbf{Effects of conformal correction and online adaptation.}
To disentangle the effects of conformal correction and online miscoverage adaptation, we conduct an ablation study based on TeaCache with \(\delta=0.08\).
We first compare TeaCache with our full CCache-enhanced version under comparable acceleration.
Since conformal correction makes cache reuse more conservative under the same threshold, we adjust the cache threshold of CCache to \(\delta=0.13\) and initialize the miscoverage level as \(\alpha_0=0.05\), yielding the same \(2.07\times\) speedup and 46s latency as TeaCache.
This setting evaluates whether the calibrated correction improves generation quality when the practical acceleration is matched.
To isolate the effect of online adaptation, we further keep the same CCache threshold \(\delta=0.13\) and compare the full adaptive version with a fixed-miscoverage variant, where the coverage level is fixed at \(1-\alpha=95\%\) throughout inference.
Since both variants use the same threshold and the same conformal correction mechanism, their latency difference mainly reflects the effect of online miscoverage adaptation.
The results in Table~\ref{tab:t2v} show that, under matched acceleration, CCache yields
performance improvements over the baseline TeaCache, confirming the effectiveness of conformal correction for safer cache reuse.
Moreover, the fixed-miscoverage variant achieves a similar average quality but is substantially slower, with speedup dropping from \(2.07\times\) to \(1.72\times\).
This verifies that online miscoverage adaptation effectively mitigates excessive conservatism and recovers acceleration without sacrificing the quality benefit of conformal calibration.

\textbf{Distributional effects of conformal correction and online adaptation.}
To further understand the behavior of CCache, we visualize the distribution of the signed oracle-to-proxy gap
\(\mathcal{E}_t=\Delta_t^{\mathrm{ora}}-\widetilde{\Delta}^{\mathrm{proxy}}_{t}\) in Figure~\ref{fig:gap_distribution_ablation}.
%
%
Figure~\ref{fig:gap_distribution_ablation}(a) compares TeaCache with its CCache-calibrated variant. The original TeaCache exhibits a positive long tail, indicating substantial proxy underestimation of the oracle discrepancy. Applying CCache shortens this unsafe tail and concentrates the distribution near the decision boundary, confirming that conformal correction targets unsafe underestimation rather than minimizing two-sided error uniformly.
Figure~\ref{fig:gap_distribution_ablation}(b) compares fixed high-coverage correction with full CCache. The fixed variant is conservative, producing more negative gaps and frequent cache refreshes. Online adaptation relaxes correction when segments are reliable, thus achieving similar quality with faster execution.

\begin{table}[t]
    \centering
    \renewcommand{\arraystretch}{1} 
    \caption{Ablation study on conformal correction and online adaptation. We compare TeaCache, full CCache under matched acceleration, and a fixed-miscoverage variant under the same CCache threshold. Better quality and lower latency are highlighted in bold.
    }
    \label{tab:t2v}

    \scalebox{0.87}{
    \begin{tabular}{l l l| c c| c c c c c c}
        \toprule
        \multirow{2}{*}{\textbf{Method}} & \multirow{2}{*}{\textbf{$\delta$}}& \multirow{2}{*}{\textbf{$\alpha$}}&\multirow{2}{*}{\textbf{Speedup}$\uparrow$} & \multirow{2}{*}{\textbf{Latency}$\downarrow$}&\multicolumn{6}{c}{\textbf{VMBench }} \tabularnewline
 &&&&&\textbf{PAS} &\textbf{OIS} &\textbf{TCS} & \textbf{MSS} & \textbf{CAS} & \textbf{Avg.}
 \\
        \midrule
Vanilla& -   &- &1.00$\times$ &  95s    & -                  & -                  & -                  &           &              & - \\
   TeaCache     & 0.08  &-&\textbf{2.07$\times$} & \textbf{46s }      & 8.78 & 67.91             & 94.08              & 71.62          & 56.44            & 59.77 \\
        
     + CCache   & 0.13   &    0.05          &\textbf{2.07$\times$} & \textbf{46s}       & \textbf{8.79} & \textbf{68.65 }            & 94.41              & 72.54          & 57.23            & \textbf{60.32} \\
  + CCache (Fixed)      & 0.13   &   0.05            &1.72$\times$ & 55s       & 8.62 & 68.26             & \textbf{94.49}              & \textbf{72.94}          & \textbf{57.29}            & \textbf{60.32} \\

        \bottomrule
    \end{tabular}
    }
    \vspace{-0.5em}
\end{table}

\begin{figure}[t]
\centering

\setlength{\tabcolsep}{4pt} 
\renewcommand{\arraystretch}{1.0} 

\begin{tabular}{cc}
    \includegraphics[width=0.485\linewidth]{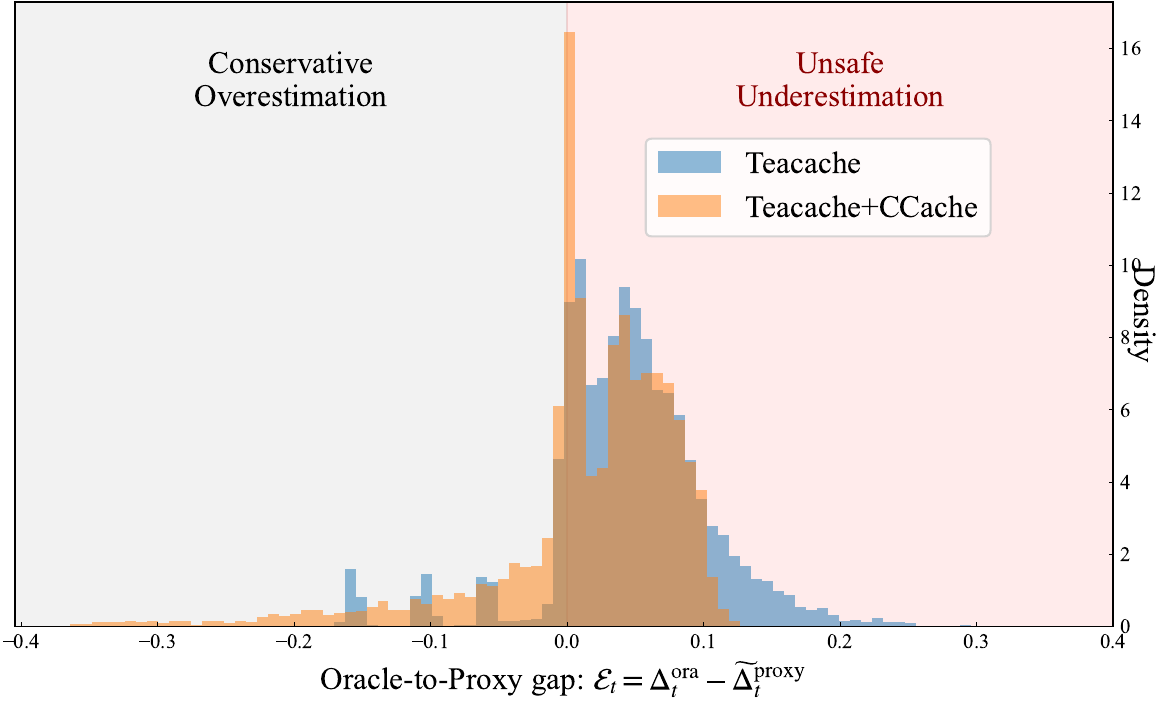} &
    \includegraphics[width=0.485\linewidth]{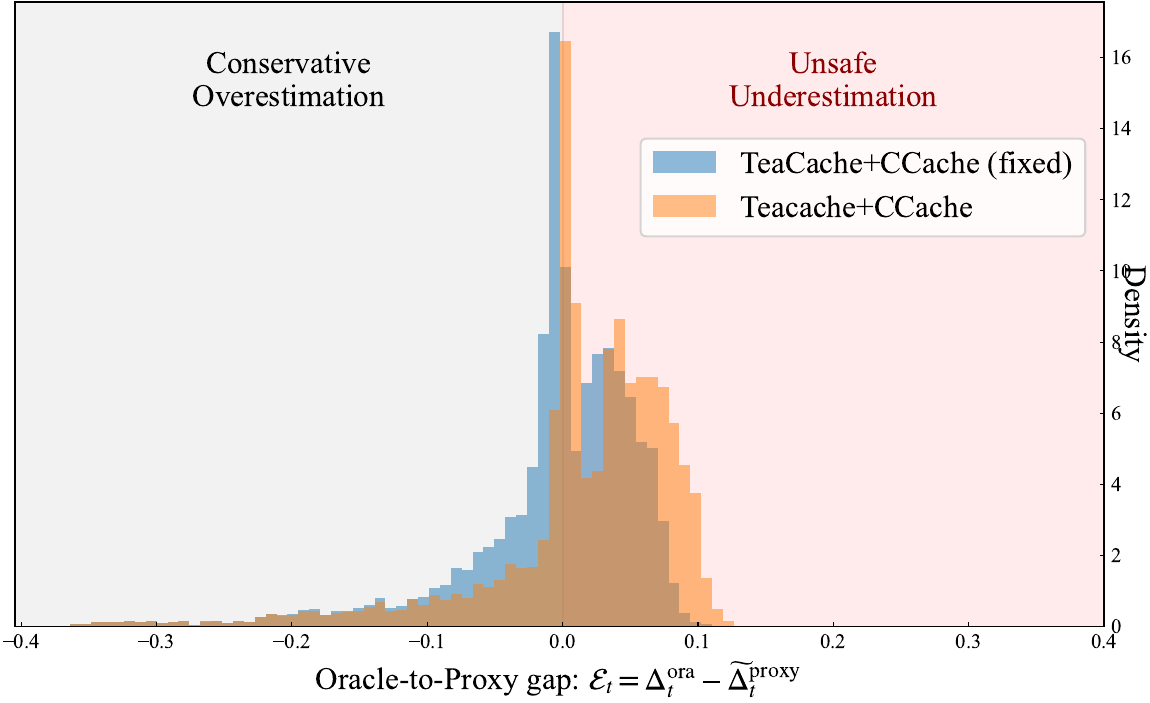} \\
    (a) TeaCache vs. CCache & (b) Fixed vs. adaptive miscoverage \\
    
\end{tabular}

\vspace{-0.5em}
\caption{Distributional analysis of conformal correction and online adaptation.}
\label{fig:gap_distribution_ablation}
\vspace{-1.5em}
\end{figure}

\textbf{Sensitivity to Calibration Set Size.}
We vary the number of calibration prompts from $N=25$ to $N=100$, while maintaining comparable acceleration. As shown in Table \ref{tab:ccache_ablation}, CCache already provides a substantial improvement with only $25$ calibration prompts. Performance then improves gradually and remains stable as the calibration set grows. These results indicate that CCache does not require a large calibration set and is relatively insensitive to the precise calibration-set size.

\begin{table}[t]
\centering
\caption{Performance comparison with different calibration-set sizes.}
\vspace{-.5em}
\label{tab:ccache_ablation}
\begin{tabular}{lcccccccc}
\toprule
\textbf{Method} & \textbf{Speedup} & \textbf{Latency} &
\textbf{PAS} & \textbf{OIS} & \textbf{TCS} &
\textbf{MSS} & \textbf{CAS} & \textbf{Avg.} \\
\midrule
DiCache ($N=0$)
& $1.76\times$ & 54s & 13.46 & 66.71 & 94.42 & 72.16 & 52.87 & 59.92 \\

+CCache ($N=25$)
& $1.70\times$ & 56s & 13.50 & 66.87 & 94.61 & 75.17 & 54.72 & 60.97 \\

+CCache ($N=50$)
& $1.73\times$ & 55s & 13.57 & 67.12 & 94.56 & 75.43 & 54.51 & 61.04 \\

+CCache ($N=75$)
& $1.76\times$ & 54s & 13.86 & 67.18 & 94.53 & 75.39 & 54.46 & 61.08 \\

+CCache ($N=100$)
& $1.73\times$ & 55s & 14.19 & 67.07 & 94.63 & 75.63 & 54.48 & 61.20 \\
\bottomrule
\end{tabular}
\vspace{-1em}
\end{table}

\textbf{OOD Transferability of Conformal Corrections.}
We further calibrate CCache using 100 prompts from a separate benchmark VBench rather than from the target VMBench evaluation pool. As shown in Table \ref{tab:ccache_comparison}, despite this prompt-distribution shift, the transferred correction achieves an average VMBench score of $61.18$, which is nearly identical to the $61.20$ obtained using in-domain VMBench calibration prompts. We hypothesize that this transferability partly benefits from the shared semantic representation induced by the same pretrained text encoder.

\begin{table}[h]
\centering
\vspace{-.75em}
\caption{Comparison of DiCache with out-of-distribution calibration set.}
\vspace{-.5em}
\label{tab:ccache_comparison}
\begin{tabular}{lcccccccc}
\toprule
\textbf{Method} & \textbf{Speedup} & \textbf{Latency} &
\textbf{PAS} & \textbf{OIS} & \textbf{TCS} &
\textbf{MSS} & \textbf{CAS} & \textbf{Avg.} \\
\midrule
DiCache
& $1.76\times$ & 54s & 13.46 & 66.71 & 94.42 & 72.16 & 52.87 & 59.92 \\

+CCache (VMBench)
& $1.73\times$ & 55s & 14.19 & 67.07 & 94.63 & 75.63 & 54.48 & 61.20 \\

+CCache (VBench)
& $1.79\times$ & 53s & 14.15 & 66.95 & 94.68 & 74.87 & 55.26 & 61.18 \\
\bottomrule
\end{tabular}
\vspace{-.75em}
\end{table}


\textbf{Latency--Quality Trade-off across Cache Thresholds}
We provide a quantitative evaluation across a broader range of caching aggressiveness levels in Table~\ref{tab:pareto_sweep}. We sweep the cache threshold $\delta$ for DiCache and CCache on Wan2.1-1.3B. CCache consistently achieves higher VMBench scores at comparable acceleration, with the advantage increasing under more aggressive caching.

\begin{table}[h]
\centering
\vspace{-.5em}
\caption{Latency--quality comparison between DiCache and CCache across acceleration levels.}
\vspace{-.5em}
\label{tab:pareto_sweep}
\setlength{\tabcolsep}{8pt}
\renewcommand{\arraystretch}{1.08}
\begin{tabular}{c|ccc|ccc}
\toprule
\multirow{2}{*}{\textbf{Speedup Level}} 
& \multicolumn{3}{c|}{\textbf{DiCache}} 
& \multicolumn{3}{c}{\textbf{+ CCache}} \\
& \(\boldsymbol{\delta}\) 
& \textbf{Speedup} 
& \textbf{Avg.} 
& \(\boldsymbol{\delta}\) 
& \textbf{Speedup} 
& \textbf{Avg.} \\
\midrule
\(\approx 1.50\times\) 
& 0.12 & \(1.50\times\) & 61.21
& 0.22 & \(1.54\times\) & 61.59 \\

\(\approx 1.75\times\) 
& 0.20 & \(1.76\times\) & 59.93
& 0.35 & \(1.73\times\) & 61.20 \\

\(\approx 1.90\times\) 
& 0.45 & \(1.89\times\) & 55.68
& 0.50 & \(1.88\times\) & 60.03 \\

\(\approx 2.00\times\) 
& 0.55 & \(1.97\times\) & 54.11
& 0.70 & \(1.95\times\) & 57.98 \\
\bottomrule
\end{tabular}
\vspace{-.5em}
\end{table}

\section{Conclusion}
\label{sec:conclusion}

In this paper, we presented CCache, a training-free conformal calibration framework for reliable cache-based acceleration in diffusion and flow-based generative models.
Rather than designing a stronger proxy discrepancy estimator, we revisited proxy-discrepancy caching from a reliability perspective and formulated cache reuse as a one-sided risk-control problem.
By calibrating timestep-wise upper corrections for the signed proxy-to-oracle gap, CCache directly targets unsafe proxy underestimation, which is the main source of cache-induced quality degradation.
To further mitigate the conservatism introduced by fixed high-coverage correction, we introduced an online miscoverage adaptation strategy that adjusts the calibration strength according to segment-level cache reliability.
CCache is plug-and-play and can be integrated with existing proxy-discrepancy caching methods without modifying or retraining the generative backbone.
Extensive experiments on both image and video generation demonstrate that CCache improves generation quality and cache reliability while preserving favorable speed-quality trade-offs.
We hope this work encourages future research on statistically calibrated and reliability-aware acceleration mechanisms for generative model inference.

\newpage
\bibliographystyle{unsrt}
\bibliography{ref_ga}

\appendix
\newpage
\include{appendix_for_arxiv}
\include{checklist}

\end{document}

%% file: appendix_for_arxiv.tex
\newpage
\section{Limitations and broader impacts}
\textbf{Limitations.}
While CCache improves the reliability of proxy-discrepancy caching, its effectiveness still depends on the quality of the underlying proxy signal.
If the original proxy is severely misaligned with the oracle discrepancy across most timesteps, conformal calibration can reduce unsafe underestimation but may require larger corrections, potentially limiting the achievable acceleration.
In addition, the validity of split conformal prediction relies on the exchangeability between calibration and evaluation prompts.
Although we construct the calibration subset from the target evaluation prompt pool to better satisfy this assumption, substantial distribution shifts may still weaken the reliability of the calibrated correction.
Moreover, our conformal guarantee is timestep-wise and marginal, and should not be interpreted as a formal trajectory-level coverage guarantee.
We believe that stronger proxy designs, larger and more diverse calibration sets, and more advanced trajectory-level risk-control strategies could further improve the reliability and efficiency of CCache.

\textbf{Broader impacts.}
Our work aims to improve the efficiency and reliability of generative model inference.
By reducing unnecessary full network evaluations, CCache may help lower computational cost and energy consumption when deploying large image and video generation models.
However, as with other acceleration techniques for generative models, improving inference efficiency may also make it easier to produce large volumes of synthetic content, including potentially misleading, biased, or harmful media.
In addition, CCache inherits the social risks of the underlying generative models, such as biased visual outputs, unsafe content generation, and possible misuse in deepfake creation.
Addressing these risks requires complementary safeguards, including dataset auditing, content moderation, watermarking, and responsible deployment policies.

\section{Appendix}

\subsection{Additional experimental details}
\label{app:exp_details}
 
\textbf{Calibration Cost.}
CCache requires only a single standard no-cache generation run for each calibration prompt. During this run, the full-model outputs and the corresponding proxy features are recorded, allowing the oracle–proxy residuals to be computed. For Wan2.1-1.3B, collecting $N=100$ calibration trajectories takes approximately 20 minutes on 8 NVIDIA H200 GPUs. Once these trajectories have been collected, sorting the residual scores and pre-extracting the timestep-wise quantiles for all considered coverage levels takes only a few seconds. This is just a one-time offline cost.

\textbf{Tuning burden for matched-speed comparisons.}
The reported matched-speed comparisons require only limited tuning: the adaptive-update hyperparameters are fixed across all experiments, and only the cache threshold $\delta$ is adjusted to match the target runtime. Since reliability is our primary objective, we fix the target miscoverage level to $\alpha_0=0.05$, rather than tuning it separately for individual models, datasets, or baselines. We use $N=100$ calibration samples, yielding an empirical quantile resolution of approximately $1/(N+1)\approx0.01$. Accordingly, we set $\eta=0.2$, with $\alpha_{min}=0.05$ and $\alpha_{max}=0.5$ throughout all experiments. Under the proposed update rule, each non-violating segment increases the miscoverage level by $0.01$, corresponding approximately to one empirical quantile level, whereas a violating segment decreases it by $0.19$. Therefore, the correction is relaxed at a practically meaningful resolution but rapidly returns toward the conservative 95\% coverage baseline once a violation is observed.

The only parameter adjusted for the matched-speed comparisons is the cache threshold $\delta$. Increasing $\delta$ generally permits longer cache reuse and therefore increases the speedup, making runtime matching a one-dimensional search problem. Because cache refreshes are discrete threshold-crossing decisions, the exact latency may change in a piecewise manner rather than perfectly smoothly; nevertheless, the operating point can be localized reliably from the calibration statistics. **Specifically, we initialize the CCache threshold by increasing the baseline threshold by approximately $(\mathrm{Speedup}-1)\bar q_{0.95}$, where $\bar q_{0.95}$ is the mean timestep-wise $95\%$-quantile correction.** The term $Speedup-1$ approximates the expected number of cache-reuse steps per full model evaluation, so this initialization compensates for the additional conformal corrections accumulated during cache reuse. We then evaluate only one or two nearby threshold values on a small fixed pilot set, five prompts in our experiments, and select the value whose latency most closely matches that of the corresponding baseline.

An illustrative example is shown below:

\begin{table}[ht]
\centering
\vspace{-.5em}
\caption{Illustration of cache-threshold adjustment for matched-speed comparison.}
\label{tab:threshold_matching}
\vspace{-.5em}
\setlength{\tabcolsep}{3pt}
\renewcommand{\arraystretch}{1.05}
\begin{tabular}{l c c c|l c c c|c c}
\toprule
\textbf{Method} & \(\boldsymbol{\delta}\) & \textbf{Speedup} & \textbf{Latency} &
\textbf{Method} & \(\boldsymbol{\delta}\) & \textbf{Speedup} & \textbf{Latency} &
\(\boldsymbol{\bar{q}_{0.95}}\) & \textbf{Estimated } \(\boldsymbol{\delta}\) \\
\midrule
TeaCache & 0.08 & \(2.07\times\) & 46s &
+ CCache & 0.13 & \(2.07\times\) & 46s &
0.028 & \(\approx 0.11\) \\

TeaCache & 0.20 & \(3.52\times\) & 27s &
+ CCache & 0.25 & \(3.39\times\) & 28s &
0.028 & \(\approx 0.27\) \\
\bottomrule
\end{tabular}
\vspace{-1em}
\end{table}

\textbf{Transferability across resolutions and aspect ratios.} Base on the T2I generation task and FLUX model, we apply the timestep-wise quantile corrections calibrated at the default $1024\times1024$ resolution directly to all seven resolution configurations, ranging from $512\times512$ to $1536\times640$, without any resolution-specific recalibration.

\begin{table}[ht]
\centering
\vspace{-.5em}
\caption{Quality and efficiency comparison across resolutions.}
\label{tab:cache_comparison}
\small
\vspace{-.5em}
\setlength{\tabcolsep}{2.5pt}
\begin{tabular}{lccccccccl}
\toprule
& \multicolumn{4}{c}{HPSv3 $\uparrow$}
& \multicolumn{3}{c}{Latency (s) $\downarrow$}
& \multicolumn{2}{c}{Speedup $\uparrow$} \\
\cmidrule(lr){2-5}
\cmidrule(lr){6-8}
\cmidrule(lr){9-10}
Resolution
& No-cache & DiCache & +CCache & $\Delta$
& No-cache & DiCache & +CCache
& DiCache & +CCache \\
\midrule

$512\times 512$
& 10.25 & 9.87 & \textbf{10.17} & \textbf{+0.30}
& 2.39 & 1.07 & \textbf{0.94}
& $2.24\times$ & \textbf{$2.54\times$} \\

$768\times 768$
& 10.35 & 10.51 & \textbf{10.75} & \textbf{+0.25}
& 3.99 & 1.67 & \textbf{1.42}
& $2.40\times$ & \textbf{$2.81\times$} \\

$1024\times 1024$
& 10.61 & 10.37 & \textbf{10.84} & \textbf{+0.46}
& 5.66 & \textbf{2.29} & 2.32
& \textbf{$2.47\times$} & $2.44\times$ \\

$1216{\times}832$
& 10.85 & 10.21 & \textbf{10.61} & \textbf{+0.40}
& 6.82 & 2.74 & \textbf{2.25}
& $2.49\times$ & \textbf{$3.03\times$} \\

$1152{\times}896$
& 10.75 & 10.21 & \textbf{10.62} & \textbf{+0.40}
& 8.30 & 2.77 & \textbf{2.28}
& $3.00\times$ & \textbf{$3.63\times$} \\

$1344{\times}768$
& 11.01 & 10.34 & \textbf{10.72} & \textbf{+0.38}
& 6.95 & 2.76 & \textbf{2.28}
& $2.52\times$ & \textbf{$3.04\times$} \\

$1536{\times}640$
& 11.06 & 10.49 & \textbf{10.85} & \textbf{+0.36}
& 8.84 & 2.64 & \textbf{2.19}
& $3.35\times$ & \textbf{$4.03\times$} \\

\bottomrule
\end{tabular}
\vspace{-1em}
\end{table}

As presented in Table~\ref{tab:cache_comparison}, CCache consistently improves HPSv3 over DiCache across all tested resolutions and aspect ratios, with gains ranging from +0.25 to +0.46. These results indicate that the calibrated residual corrections are robust to substantial changes in spatial resolution and aspect ratio.
In contrast, we do not recommend directly transferring the same correction vector across substantially different models, tasks, samplers, or sampling schedules. The validity of the calibrated correction relies on approximate exchangeability between the calibration and evaluation nonconformity scores. Such changes may alter the distribution of the proxy-to-oracle residuals and therefore violate the exchangeability assumption underlying the conformal calibration. Moreover, CCache uses timestep-indexed corrections, and different generation settings may have different trajectory lengths and timestep semantics.
\subsection{Additional reliability analysis}
\label{app:reliability}

\textbf{Timestep- and Segment-Level Reliability}
Although the formal conformal guarantee of CCache is timestep-wise and marginal, this reflects a fundamental theory–practice trade-off in adaptive cache-based inference. A trajectory-level statement is substantially more challenging because the segment length is determined by a threshold-crossing stopping rule, discrepancies are dependent across timesteps, and cache reuse itself may alter the subsequent latent trajectory. Our current analysis therefore makes the more limited claim that, at each timestep, the conformal correction controls the one-sided proxy-underestimation risk under the stated exchangeability assumption.

Moreover, the refresh action is a discrete threshold decision. Small residual estimation errors that do not move the accumulated discrepancy across the refresh threshold leave the resulting cache schedule unchanged. This provides a degree of practical robustness away from the decision boundary.

We wish to emphasize that despite this marginal guarantee, our proposed CCache remains a powerful and generally applicable contribution. The consistent improvements across caching baselines, model scales, and image and video generation tasks provide empirical evidence that timestep-wise one-sided calibration is practically effective. This indicates that while the conformal guarantee is timestep-wise and marginal, the overall value and utility of our framework across different model scales and generation tasks are robust.

To directly quantify the practical risk, we conduct an oracle-instrumented on-policy evaluation on 100 held-out prompts using Wan2.1-T2V-1.3B in Table~\ref{tab:empirical_reliability}. We preserve the exact cached trajectory and refresh schedule, while additionally running the full Transformer at every timestep solely for measurement. For each skipped timestep, we record the adjacent-step oracle discrepancy and accumulate it within the current cache segment. A segment violation is counted only when this discrepancy exceeds $\delta=0.13$ before the actual refresh, meaning that the cache is genuinely reused after the oracle criterion has been violated.

\begin{table}[ht]
\centering
\vspace{-.5em}
\caption{Empirical reliability of CCache under on-policy cached inference.}
\label{tab:empirical_reliability}
\setlength{\tabcolsep}{6pt}
\renewcommand{\arraystretch}{1.08}
\begin{tabular}{l c}
\toprule
\textbf{Empirical reliability metric} & \textbf{Result} \\
\midrule
Evaluated skipped Cond/Uncond passes & 5,140 \\
Passes satisfying the upper bound & 4,580 \\
Fixed-\(q_{0.95}\) coverage on skipped passes & 89.11\% \\
Evaluated cache segments & 1,366 \\
Segment validity rate & 93.70\% \\
Unsafe reuse steps per violating segment & 1.00 \\
\bottomrule
\end{tabular}
\vspace{-1em}
\end{table}

For the timestep-level diagnostic, 4,580 of the 5,140 skipped Cond/Uncond passes satisfy$\Delta^{\mathrm{ora}}_{t} \le \Delta^{\mathrm{proxy}}_{t}+q_{t,0.95}$, yielding a fixed-$q_{0.95}$ conditional coverage of 89.11\%. At the segment level, 93.70\% of cache segments satisfy the cumulative oracle criterion throughout reuse. Moreover, when a violation occurs, the method refreshes after only one additional reuse step on average, indicating that the severity of the observed violations is limited. Together with the distributional analysis in Fig. 3, which shows a substantial reduction in the unsafe positive residual tail, this experiment provides complementary evidence at both the timestep and segment levels.

\textbf{Trajectory Shift under Cached Inference.}
Calibration is performed on no-cache trajectories, whereas cache reuse may perturb the latent state encountered during accelerated inference. Consequently, the residual distribution encountered during accelerated inference may differ from that observed during no-cache calibration. 

In practice, however, this shift is mitigated by the periodic refresh mechanism. Cache reuse is restricted to relatively short segments under the local temporal redundancy assumption. At each refresh timestep, the model performs a full network evaluation of the noise/velocity prediction at the current accelerated latent state and resets the cache anchor. This re-anchors the cached approximation to the current on-policy trajectory and prevents errors from an increasingly stale model output from accumulating indefinitely.

CCache further reacts to residual shift through its online adaptation. At each refresh point, full-model outputs are available for evaluating the reliability of the preceding cache segment. When the observed discrepancy exceeds the tolerance, CCache decreases the miscoverage level and applies a more conservative correction to subsequent segments; otherwise, it gradually relaxes the correction. Thus, although the initial quantiles are obtained from no-cache trajectories, the online controller can partially compensate when the residual behavior encountered during accelerated inference becomes less reliable. 

The empirical results suggest that this periodic re-anchoring and adaptation are effective in practice. In particular, CCache continues to improve generation quality under aggressive caching settings, where cache-induced trajectory deviation is expected to be more pronounced. The distributional analysis in Figure~\ref{fig:gap_distribution_ablation} also shows that CCache substantially reduces the positive residual tail associated with unsafe underestimation.







\subsection{Additional qualitative results}
\label{appd:more_qualitative_results}
\begin{figure}[ht]
\footnotesize
\centering
\vspace{-1em}
\includegraphics[width=1\linewidth]{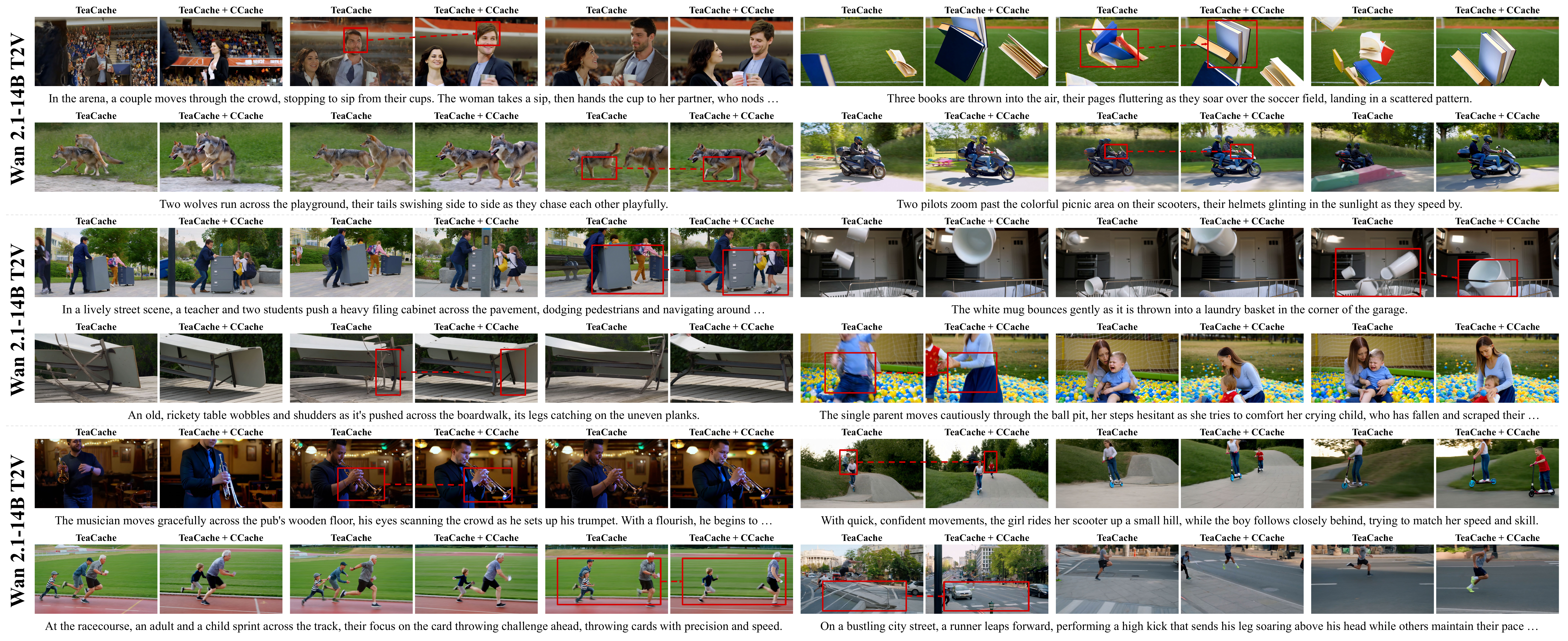}
\vspace{-2em}
\caption{
Additional qualitative comparisons of TeaCache with and without CCache on Wan2.1-T2V-14B. Red boxes highlight representative differences.
}
\label{fig:add_qualitative_comparison_wan2.1-1.3}
\vspace{-1em}
\end{figure}

\begin{figure}[ht]
\footnotesize
\centering
\includegraphics[width=1\linewidth]{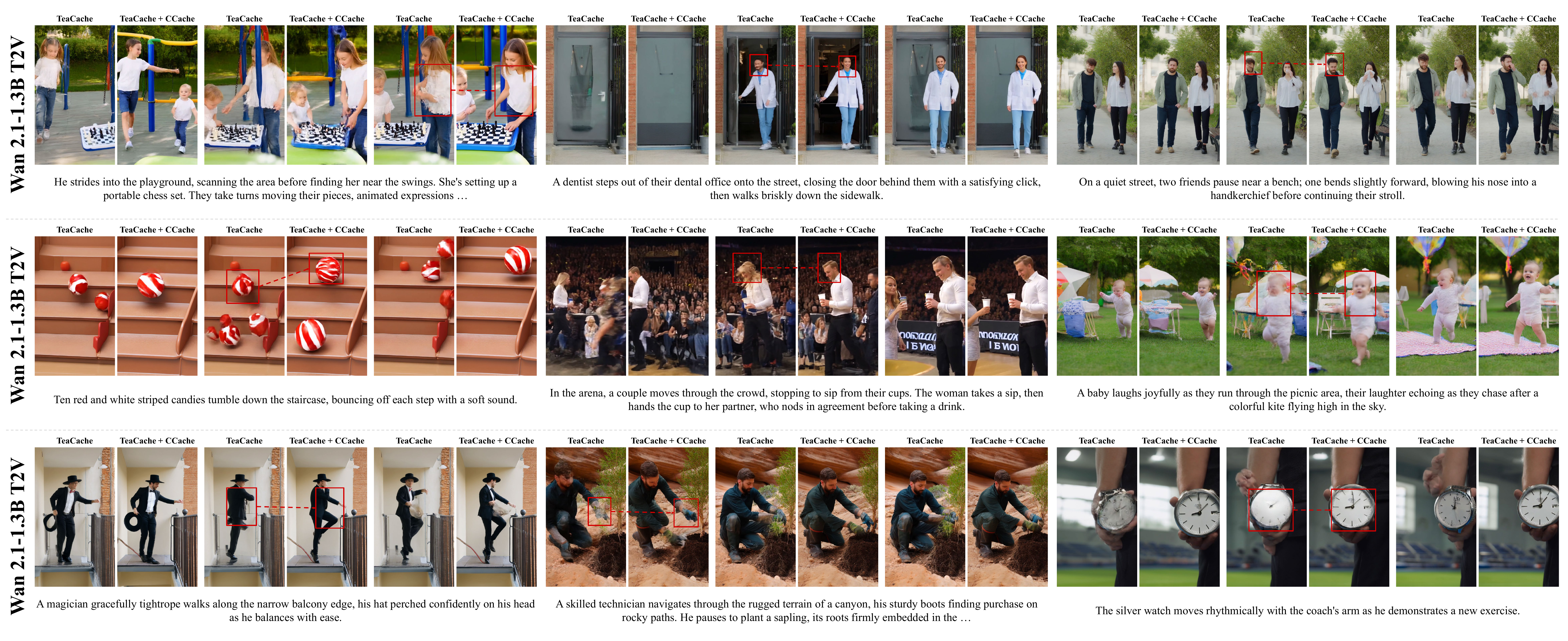}
\vspace{-2em}
\caption{
Additional qualitative comparisons of TeaCache with and without CCache on Wan2.1-T2V-1.3B. Red boxes highlight representative differences.
}
\label{fig:add_qualitative_comparison_wan2.1-14}
\vspace{-1em}
\end{figure}

We provide additional qualitative results to complement the evaluation in Figure~\ref{fig:add_qualitative_comparison_wan2.1-1.3} and Figure~\ref{fig:add_qualitative_comparison_wan2.1-14}, comparing caching baselines with and without CCache across diverse scenes and generation settings. The examples include human interactions, animal motion, and objects with intricate structures, illustrating improvements in subject appearance, object geometry, and fine visual details. For video generation, we present multiple frames sampled at identical temporal indices for both methods, allowing direct comparison at different stages of each sequence. The selected examples show that CCache can alleviate ghosting, structural distortions, and unintended duplication, and in some cases improve alignment with the prompt. Corresponding prompts and highlighted regions facilitate inspection of these differences.

\subsection{Additional quality evaluation}
\label{app:quality_metrics}

\textbf{Full-Reference Fidelity and Perceptual Metrics}
To complement HPSv3 and VMBench, we conduct additional evaluations using both full-reference fidelity metrics and reference-free perceptual and text-alignment metrics.
For the full-reference evaluation, we treat the output of the original no-cache model as the reference and compare it with the accelerated output. We report LPIPS, SSIM, and PSNR, which directly measure how faithfully each caching method preserves the output of the underlying generative model. As presented in Table~\ref{tab:additional_metrics}, we additionally report ImageReward and CLIPScore to evaluate perceptual preference and prompt–image alignment, respectively.

\begin{table}[ht]
\centering
\vspace{-.5em}
\caption{Additional fidelity and perceptual quality evaluation for TeaCache and its CCache-enhanced variant.}
\vspace{-0.5em}
\label{tab:additional_metrics}
\setlength{\tabcolsep}{5pt}
\renewcommand{\arraystretch}{1.08}
\begin{tabular}{l|cccccc}
\toprule
\textbf{Method} 
& \textbf{LPIPS}$\downarrow$
& \textbf{SSIM}$\uparrow$
& \textbf{PSNR}$\uparrow$
& \textbf{HPSv3}$\uparrow$
& \textbf{ImageReward}$\uparrow$
& \textbf{CLIPScore}$\uparrow$ \\
\midrule
TeaCache (\(\delta=0.40\))
& 0.2715
& 0.7680
& 20.11
& 9.99
& 0.7953
& 0.8024 \\

+ CCache
& \textbf{0.1694}
& \textbf{0.8462}
& \textbf{24.03}
& \textbf{10.13}
& \textbf{0.8027}
& \textbf{0.8026} \\
\midrule
TeaCache (\(\delta=0.80\))
& 0.4370
& 0.6649
& 16.89
& 8.38
& 0.5819
& 0.7930 \\

+ CCache
& \textbf{0.3593}
& \textbf{0.7267}
& \textbf{19.89}
& \textbf{9.54}
& \textbf{0.6996}
& \textbf{0.7999} \\
\bottomrule
\end{tabular}
\end{table}

%% file: checklist.tex
\section*{NeurIPS Paper Checklist}

\begin{enumerate}

\item {\bf Claims}
    \item[] Question: Do the main claims made in the abstract and introduction accurately reflect the paper's contributions and scope?
    \item[] Answer: \answerYes{} 
    \item[] Justification: The abstract and introduction accurately summarize our main theoretical findings and empirical results.
    \item[] Guidelines:
    \begin{itemize}
        \item The answer \answerNA{} means that the abstract and introduction do not include the claims made in the paper.
        \item The abstract and/or introduction should clearly state the claims made, including the contributions made in the paper and important assumptions and limitations. A \answerNo{} or \answerNA{} answer to this question will not be perceived well by the reviewers. 
        \item The claims made should match theoretical and experimental results, and reflect how much the results can be expected to generalize to other settings. 
        \item It is fine to include aspirational goals as motivation as long as it is clear that these goals are not attained by the paper. 
    \end{itemize}

\item {\bf Limitations}
    \item[] Question: Does the paper discuss the limitations of the work performed by the authors?
    \item[] Answer: \answerYes{} 
    \item[] Justification: We have discuss limitations in Appendix.
    \item[] Guidelines:
    \begin{itemize}
        \item The answer \answerNA{} means that the paper has no limitation while the answer \answerNo{} means that the paper has limitations, but those are not discussed in the paper. 
        \item The authors are encouraged to create a separate ``Limitations'' section in their paper.
        \item The paper should point out any strong assumptions and how robust the results are to violations of these assumptions (e.g., independence assumptions, noiseless settings, model well-specification, asymptotic approximations only holding locally). The authors should reflect on how these assumptions might be violated in practice and what the implications would be.
        \item The authors should reflect on the scope of the claims made, e.g., if the approach was only tested on a few datasets or with a few runs. In general, empirical results often depend on implicit assumptions, which should be articulated.
        \item The authors should reflect on the factors that influence the performance of the approach. For example, a facial recognition algorithm may perform poorly when image resolution is low or images are taken in low lighting. Or a speech-to-text system might not be used reliably to provide closed captions for online lectures because it fails to handle technical jargon.
        \item The authors should discuss the computational efficiency of the proposed algorithms and how they scale with dataset size.
        \item If applicable, the authors should discuss possible limitations of their approach to address problems of privacy and fairness.
        \item While the authors might fear that complete honesty about limitations might be used by reviewers as grounds for rejection, a worse outcome might be that reviewers discover limitations that aren't acknowledged in the paper. The authors should use their best judgment and recognize that individual actions in favor of transparency play an important role in developing norms that preserve the integrity of the community. Reviewers will be specifically instructed to not penalize honesty concerning limitations.
    \end{itemize}

\item {\bf Theory assumptions and proofs}
    \item[] Question: For each theoretical result, does the paper provide the full set of assumptions and a complete (and correct) proof?
    \item[] Answer: \answerNA{} 
    \item[] Justification: We do not have theoretical results.
    \item[] Guidelines:
    \begin{itemize}
        \item The answer \answerNA{} means that the paper does not include theoretical results. 
        \item All the theorems, formulas, and proofs in the paper should be numbered and cross-referenced.
        \item All assumptions should be clearly stated or referenced in the statement of any theorems.
        \item The proofs can either appear in the main paper or the supplemental material, but if they appear in the supplemental material, the authors are encouraged to provide a short proof sketch to provide intuition. 
        \item Inversely, any informal proof provided in the core of the paper should be complemented by formal proofs provided in appendix or supplemental material.
        \item Theorems and Lemmas that the proof relies upon should be properly referenced. 
    \end{itemize}

    \item {\bf Experimental result reproducibility}
    \item[] Question: Does the paper fully disclose all the information needed to reproduce the main experimental results of the paper to the extent that it affects the main claims and/or conclusions of the paper (regardless of whether the code and data are provided or not)?
    \item[] Answer: \answerYes{} 
    \item[] Justification: We provide all necessary details for reproduction, including hyperparameters, model architectures, and dataset configurations.
    \item[] Guidelines:
    \begin{itemize}
        \item The answer \answerNA{} means that the paper does not include experiments.
        \item If the paper includes experiments, a \answerNo{} answer to this question will not be perceived well by the reviewers: Making the paper reproducible is important, regardless of whether the code and data are provided or not.
        \item If the contribution is a dataset and\slash or model, the authors should describe the steps taken to make their results reproducible or verifiable. 
        \item Depending on the contribution, reproducibility can be accomplished in various ways. For example, if the contribution is a novel architecture, describing the architecture fully might suffice, or if the contribution is a specific model and empirical evaluation, it may be necessary to either make it possible for others to replicate the model with the same dataset, or provide access to the model. In general. releasing code and data is often one good way to accomplish this, but reproducibility can also be provided via detailed instructions for how to replicate the results, access to a hosted model (e.g., in the case of a large language model), releasing of a model checkpoint, or other means that are appropriate to the research performed.
        \item While NeurIPS does not require releasing code, the conference does require all submissions to provide some reasonable avenue for reproducibility, which may depend on the nature of the contribution. For example
        \begin{enumerate}
            \item If the contribution is primarily a new algorithm, the paper should make it clear how to reproduce that algorithm.
            \item If the contribution is primarily a new model architecture, the paper should describe the architecture clearly and fully.
            \item If the contribution is a new model (e.g., a large language model), then there should either be a way to access this model for reproducing the results or a way to reproduce the model (e.g., with an open-source dataset or instructions for how to construct the dataset).
            \item We recognize that reproducibility may be tricky in some cases, in which case authors are welcome to describe the particular way they provide for reproducibility. In the case of closed-source models, it may be that access to the model is limited in some way (e.g., to registered users), but it should be possible for other researchers to have some path to reproducing or verifying the results.
        \end{enumerate}
    \end{itemize}

\item {\bf Open access to data and code}
    \item[] Question: Does the paper provide open access to the data and code, with sufficient instructions to faithfully reproduce the main experimental results, as described in supplemental material?
    \item[] Answer: \answerYes{} 
    \item[] Justification: All open-source models and datasets used in our experiments are properly cited.
    \item[] Guidelines:
    \begin{itemize}
        \item The answer \answerNA{} means that paper does not include experiments requiring code.
        \item Please see the NeurIPS code and data submission guidelines (\url{https://neurips.cc/public/guides/CodeSubmissionPolicy}) for more details.
        \item While we encourage the release of code and data, we understand that this might not be possible, so \answerNo{} is an acceptable answer. Papers cannot be rejected simply for not including code, unless this is central to the contribution (e.g., for a new open-source benchmark).
        \item The instructions should contain the exact command and environment needed to run to reproduce the results. See the NeurIPS code and data submission guidelines (\url{https://neurips.cc/public/guides/CodeSubmissionPolicy}) for more details.
        \item The authors should provide instructions on data access and preparation, including how to access the raw data, preprocessed data, intermediate data, and generated data, etc.
        \item The authors should provide scripts to reproduce all experimental results for the new proposed method and baselines. If only a subset of experiments are reproducible, they should state which ones are omitted from the script and why.
        \item At submission time, to preserve anonymity, the authors should release anonymized versions (if applicable).
        \item Providing as much information as possible in supplemental material (appended to the paper) is recommended, but including URLs to data and code is permitted.
    \end{itemize}

\item {\bf Experimental setting/details}
    \item[] Question: Does the paper specify all the training and test details (e.g., data splits, hyperparameters, how they were chosen, type of optimizer) necessary to understand the results?
    \item[] Answer: \answerYes{} 
    \item[] Justification: We have pecify all the training and test details (e.g., data splits, hyperparameters, how they were chosen, type of optimizer) necessary to understand the results in experimental settings.
    \item[] Guidelines:
    \begin{itemize}
        \item The answer \answerNA{} means that the paper does not include experiments.
        \item The experimental setting should be presented in the core of the paper to a level of detail that is necessary to appreciate the results and make sense of them.
        \item The full details can be provided either with the code, in appendix, or as supplemental material.
    \end{itemize}

\item {\bf Experiment statistical significance}
    \item[] Question: Does the paper report error bars suitably and correctly defined or other appropriate information about the statistical significance of the experiments?
    \item[] Answer: \answerNo{} 
    \item[] Justification: We cannot provide error bar for some competitors. Besides, the current results are sufficient to show the efficacy of our method.
    \item[] Guidelines:
    \begin{itemize}
        \item The answer \answerNA{} means that the paper does not include experiments.
        \item The authors should answer \answerYes{} if the results are accompanied by error bars, confidence intervals, or statistical significance tests, at least for the experiments that support the main claims of the paper.
        \item The factors of variability that the error bars are capturing should be clearly stated (for example, train/test split, initialization, random drawing of some parameter, or overall run with given experimental conditions).
        \item The method for calculating the error bars should be explained (closed form formula, call to a library function, bootstrap, etc.)
        \item The assumptions made should be given (e.g., Normally distributed errors).
        \item It should be clear whether the error bar is the standard deviation or the standard error of the mean.
        \item It is OK to report 1-sigma error bars, but one should state it. The authors should preferably report a 2-sigma error bar than state that they have a 96\% CI, if the hypothesis of Normality of errors is not verified.
        \item For asymmetric distributions, the authors should be careful not to show in tables or figures symmetric error bars that would yield results that are out of range (e.g., negative error rates).
        \item If error bars are reported in tables or plots, the authors should explain in the text how they were calculated and reference the corresponding figures or tables in the text.
    \end{itemize}

\item {\bf Experiments compute resources}
    \item[] Question: For each experiment, does the paper provide sufficient information on the computer resources (type of compute workers, memory, time of execution) needed to reproduce the experiments?
    \item[] Answer: \answerYes{} 
    \item[] Justification: We have provide sufficient information on the computer resources in experimental settings.
    \item[] Guidelines:
    \begin{itemize}
        \item The answer \answerNA{} means that the paper does not include experiments.
        \item The paper should indicate the type of compute workers CPU or GPU, internal cluster, or cloud provider, including relevant memory and storage.
        \item The paper should provide the amount of compute required for each of the individual experimental runs as well as estimate the total compute. 
        \item The paper should disclose whether the full research project required more compute than the experiments reported in the paper (e.g., preliminary or failed experiments that didn't make it into the paper). 
    \end{itemize}
    
\item {\bf Code of ethics}
    \item[] Question: Does the research conducted in the paper conform, in every respect, with the NeurIPS Code of Ethics \url{https://neurips.cc/public/EthicsGuidelines}?
    \item[] Answer: \answerYes{} 
    \item[] Justification: Our research focuses on general visual tasks, which totally conforms with the NeurIPS Code of Ethics.
    \item[] Guidelines:
    \begin{itemize}
        \item The answer \answerNA{} means that the authors have not reviewed the NeurIPS Code of Ethics.
        \item If the authors answer \answerNo, they should explain the special circumstances that require a deviation from the Code of Ethics.
        \item The authors should make sure to preserve anonymity (e.g., if there is a special consideration due to laws or regulations in their jurisdiction).
    \end{itemize}

\item {\bf Broader impacts}
    \item[] Question: Does the paper discuss both potential positive societal impacts and negative societal impacts of the work performed?
    \item[] Answer: \answerYes{} 
    \item[] Justification: We discuss both potential positive societal impacts and negative societal impacts of the work in the Appendix.
    \item[] Guidelines:
    \begin{itemize}
        \item The answer \answerNA{} means that there is no societal impact of the work performed.
        \item If the authors answer \answerNA{} or \answerNo, they should explain why their work has no societal impact or why the paper does not address societal impact.
        \item Examples of negative societal impacts include potential malicious or unintended uses (e.g., disinformation, generating fake profiles, surveillance), fairness considerations (e.g., deployment of technologies that could make decisions that unfairly impact specific groups), privacy considerations, and security considerations.
        \item The conference expects that many papers will be foundational research and not tied to particular applications, let alone deployments. However, if there is a direct path to any negative applications, the authors should point it out. For example, it is legitimate to point out that an improvement in the quality of generative models could be used to generate Deepfakes for disinformation. On the other hand, it is not needed to point out that a generic algorithm for optimizing neural networks could enable people to train models that generate Deepfakes faster.
        \item The authors should consider possible harms that could arise when the technology is being used as intended and functioning correctly, harms that could arise when the technology is being used as intended but gives incorrect results, and harms following from (intentional or unintentional) misuse of the technology.
        \item If there are negative societal impacts, the authors could also discuss possible mitigation strategies (e.g., gated release of models, providing defenses in addition to attacks, mechanisms for monitoring misuse, mechanisms to monitor how a system learns from feedback over time, improving the efficiency and accessibility of ML).
    \end{itemize}
    
\item {\bf Safeguards}
    \item[] Question: Does the paper describe safeguards that have been put in place for responsible release of data or models that have a high risk for misuse (e.g., pre-trained language models, image generators, or scraped datasets)?
    \item[] Answer: \answerNA{} 
    \item[] Justification: The paper poses no such risks.
    \item[] Guidelines:
    \begin{itemize}
        \item The answer \answerNA{} means that the paper poses no such risks.
        \item Released models that have a high risk for misuse or dual-use should be released with necessary safeguards to allow for controlled use of the model, for example by requiring that users adhere to usage guidelines or restrictions to access the model or implementing safety filters. 
        \item Datasets that have been scraped from the Internet could pose safety risks. The authors should describe how they avoided releasing unsafe images.
        \item We recognize that providing effective safeguards is challenging, and many papers do not require this, but we encourage authors to take this into account and make a best faith effort.
    \end{itemize}

\item {\bf Licenses for existing assets}
    \item[] Question: Are the creators or original owners of assets (e.g., code, data, models), used in the paper, properly credited and are the license and terms of use explicitly mentioned and properly respected?
    \item[] Answer: \answerYes{} 
    \item[] Justification: We cite all used public datasets and pre-trained models in Section 4.
    \item[] Guidelines:
    \begin{itemize}
        \item The answer \answerNA{} means that the paper does not use existing assets.
        \item The authors should cite the original paper that produced the code package or dataset.
        \item The authors should state which version of the asset is used and, if possible, include a URL.
        \item The name of the license (e.g., CC-BY 4.0) should be included for each asset.
        \item For scraped data from a particular source (e.g., website), the copyright and terms of service of that source should be provided.
        \item If assets are released, the license, copyright information, and terms of use in the package should be provided. For popular datasets, \url{paperswithcode.com/datasets} has curated licenses for some datasets. Their licensing guide can help determine the license of a dataset.
        \item For existing datasets that are re-packaged, both the original license and the license of the derived asset (if it has changed) should be provided.
        \item If this information is not available online, the authors are encouraged to reach out to the asset's creators.
    \end{itemize}

\item {\bf New assets}
    \item[] Question: Are new assets introduced in the paper well documented and is the documentation provided alongside the assets?
    \item[] Answer: \answerNA{} 
    \item[] Justification: We do not introduce new assets.
    \item[] Guidelines:
    \begin{itemize}
        \item The answer \answerNA{} means that the paper does not release new assets.
        \item Researchers should communicate the details of the dataset\slash code\slash model as part of their submissions via structured templates. This includes details about training, license, limitations, etc. 
        \item The paper should discuss whether and how consent was obtained from people whose asset is used.
        \item At submission time, remember to anonymize your assets (if applicable). You can either create an anonymized URL or include an anonymized zip file.
    \end{itemize}

\item {\bf Crowdsourcing and research with human subjects}
    \item[] Question: For crowdsourcing experiments and research with human subjects, does the paper include the full text of instructions given to participants and screenshots, if applicable, as well as details about compensation (if any)? 
    \item[] Answer:  \answerNA{} 
    \item[] Justification: We do not include any crowdsourcing experiments and research with human subjects in our paper.
    \item[] Guidelines:
    \begin{itemize}
        \item The answer \answerNA{} means that the paper does not involve crowdsourcing nor research with human subjects.
        \item Including this information in the supplemental material is fine, but if the main contribution of the paper involves human subjects, then as much detail as possible should be included in the main paper. 
        \item According to the NeurIPS Code of Ethics, workers involved in data collection, curation, or other labor should be paid at least the minimum wage in the country of the data collector. 
    \end{itemize}

\item {\bf Institutional review board (IRB) approvals or equivalent for research with human subjects}
    \item[] Question: Does the paper describe potential risks incurred by study participants, whether such risks were disclosed to the subjects, and whether Institutional Review Board (IRB) approvals (or an equivalent approval/review based on the requirements of your country or institution) were obtained?
    \item[] Answer:  \answerNA{} 
    \item[] Justification: We only conduct objective experiments in our paper.
    \item[] Guidelines:
    \begin{itemize}
        \item The answer \answerNA{} means that the paper does not involve crowdsourcing nor research with human subjects.
        \item Depending on the country in which research is conducted, IRB approval (or equivalent) may be required for any human subjects research. If you obtained IRB approval, you should clearly state this in the paper. 
        \item We recognize that the procedures for this may vary significantly between institutions and locations, and we expect authors to adhere to the NeurIPS Code of Ethics and the guidelines for their institution. 
        \item For initial submissions, do not include any information that would break anonymity (if applicable), such as the institution conducting the review.
    \end{itemize}

\item {\bf Declaration of LLM usage}
    \item[] Question: Does the paper describe the usage of LLMs if it is an important, original, or non-standard component of the core methods in this research? Note that if the LLM is used only for writing, editing, or formatting purposes and does \emph{not} impact the core methodology, scientific rigor, or originality of the research, declaration is not required.
    \item[] Answer:  \answerNA{} 
    \item[] Justification: This work does not incorporate Large Language Models (LLMs) into its core methodology. Consequently, no description of such usage is provided
    \item[] Guidelines:
    \begin{itemize}
        \item The answer \answerNA{} means that the core method development in this research does not involve LLMs as any important, original, or non-standard components.
        \item Please refer to our LLM policy in the NeurIPS handbook for what should or should not be described.
    \end{itemize}

\end{enumerate}